\documentclass{article}

\usepackage[main,final]{neurips_2026}

\usepackage[utf8]{inputenc} 
\usepackage[T1]{fontenc}    
\usepackage{hyperref}       
\usepackage{url}            
\usepackage{booktabs}       
\usepackage{amsfonts}       
\usepackage{nicefrac}       
\usepackage{microtype}      
\usepackage{xcolor}         
\usepackage{caption}
\usepackage{adjustbox} 
\usepackage{amsmath}
\usepackage{float} 
\usepackage{makecell}
\title{DinoComplete: 3D Shape Completion with Distilled Semantic Priors and State Space Models}

\author{%
  Furkan Mert Algan \quad  Eckehard Steinbach \\
  Chair of Media Technology\\
  Munich Institute of Robotics and Machine Intelligence\\
  School of Computation Information and Technology, Technical University of Munich\\
  \texttt{\{fmert.algan, eckehard.steinbach\}@tum.de}
}

\begin{document}

\maketitle

\begin{abstract}
3D shape completion from partial scans remains challenging for unseen categories and noisy real-world observations, where geometry alone is often insufficient for inferring missing structure. We present DinoComplete, a deterministic and efficient shape completion framework that augments geometric reconstruction with voxel-aligned semantic priors distilled from DINO features. First, we construct multi-view DINO feature volumes aligned with ShapeNet data and train a student network to predict dense semantic features directly from incomplete shapes. These predicted features capture global structure and part-aware semantic context while remaining aligned with the underlying geometry. We then integrate these distilled features into a completion network, where geometric and semantic voxel representations are fused through voxel state-space modeling. To enable efficient long-range reasoning without sacrificing resolution, we introduce a Multi-Scale Voxel Mamba module that refines the fused features by combining full-grid and chunk-wise sequence modeling. Experiments on unseen ShapeNet categories and ScanNet objects show that DinoComplete achieves competitive completion quality than prior deterministic and generative based completion methods while using relatively fewer parameters, requiring lower memory, and achieving faster inference. Our results demonstrate that distilling semantic priors from visual foundation models improves generalization and robustness in 3D shape completion.

\end{abstract}

\section{Introduction}
\label{introduction}
The availability of consumer-grade depth sensors, such as Microsoft Kinect and Intel RealSense, has led to significant progress in 3D reconstruction by enabling efficient capture of RGB-D data in real-world environments. These sensors have enabled the creation of large-scale 3D datasets~\cite{dai2017scannet,song2015sun,chang2017matterport3d} and supported applications in mixed reality, robotics, and digital content creation, with early systems demonstrating real-time dense reconstruction from streaming depth input~\cite{newcombe2011kinectfusion, niessner2013real, dai2017bundlefusion}.

Although  these methods make reconstruction simple for users, the reconstructed 3D models often suffer from incomplete geometry, as well as noise and clutter caused by occlusions. This can lead to degraded geometric quality and missing structural details, which significantly limits their usability. Current approaches to 3D completion primarily operate on representations such as point clouds, voxel grids, or multi-view images. However, despite the fact that commodity RGB-D sensors produce truncated signed distance functions (TSDFs) as an intermediate representation in  real-time reconstruction pipelines~\cite{newcombe2011kinectfusion, niessner2013real, dai2017bundlefusion}, relatively few methods directly operate on distance functions. 

TSDF representations encode geometric structure but lack high-level semantic information, making them difficult to process under noisy and partial observations. Existing approaches such as PatchComplete~\cite{tang2022patchcomplete} focus on learning local geometric priors by retrieving and assembling shape patches, but struggle with decoupled patch-wise training. Probabilistic methods, such as DiffComplete~\cite{chu2023diffcomplete}, adopt diffusion-based generative modeling to capture complex shape distributions and improve global reasoning. However, they remain limited by high computational cost.
This highlights the need for methods that integrate semantic understanding with efficient and scalable inference.

We present DinoComplete, an architecture that complements TSDF representations with rich semantic features to enable robust and globally consistent reconstruction. To the best of our knowledge, we are the first to distill DINO-based semantic features into a TSDF-only shape completion pipeline, enabling semantic reasoning directly from partial geometry without requiring additional information at inference time. We distill semantic information from incomplete shapes into a 3D student–teacher model, which produces voxel-aligned feature volumes capturing global semantic context. To effectively model long-range dependencies in 3D, we introduce a voxel-based state space model that enables efficient global reasoning over volumetric features. Building on these components, our full architecture integrates semantic and geometric representations to reconstruct complete shapes from noisy and partial inputs.

In summary, our contributions are as follows.

\begin{itemize}
    \item A 3D student--teacher distillation architecture that learns voxel-aligned semantic features from incomplete TSDF inputs, providing strong global priors for shape understanding
    
    \item A chunk-based voxel state space model for efficient global context aggregation in 3D, enabling scalable modeling of long-range dependencies
    
    \item A shape completion framework integrating distilled semantic features with global voxel reasoning, achieving the best average results on seen and unseen ShapeNet categories and competitive performance on real-world ScanNet objects
\end{itemize}

\section{Related Work}
\label{related_work}

\textbf{Semantic feature learning.}
Self-supervised visual foundation models~\citep{he2020moco,chen2020simclr,grill2020byol,oquab2023dinov2,simeoni2025dinov3} have become a powerful source of semantic priors for geometric tasks.
DINO~\citep{caron2021dino} learns image representations via self-distillation, where a student network is trained to match the output of a momentum teacher under different image augmentations.
This process produces dense feature embeddings for each image patch that capture high-level semantic structure, such as object parts, boundaries, and category-level similarity, without requiring manual labels.
DINOv2~\citep{oquab2023dinov2} scales this paradigm to larger datasets and architectures, improving robustness and transferability across domains, while DINOv3~\citep{simeoni2025dinov3} further enhances feature quality through large-scale training and improved dense supervision, enabling more precise structured predictions.

Recent works have used DINO features as a representation across a wide range of downstream tasks such as indoor perception ~\citep{knaebel2025dino} and vision-language-action models.~\citep{kim2024openvla}. 
For geometric reasoning, DINO features have been applied to tasks such as visual odometry~\citep{azhari2025dino} and monocular 3D estimation~\citep{hu2024metric3dv2, wang2025moge, wang2025mogev2}, demonstrating strong robustness under challenging visual conditions. 
Building on these advances, recent works explore the use of DINO features for 3D-aware modeling and reconstruction, including object-centric pipelines such as AutoRecon~\citep{wang2023autorecon}. 
Most notably, geometry-aware architectures such as VGGT~\citep{wang2025vggt} show that DINO-style representations can directly support 3D reasoning, including depth, camera estimation, and spatial structure. 
These works demonstrate that DINO features support 3D reconstruction and suggest their potential for integration with volumetric representations such as TSDF-based modeling.

\textbf{Shape completion.}
3D reconstruction has been widely studied across different representations, including point clouds~\citep{adapointr, yuan2018pcn, yu2021pointr} and multi-view images~\citep{schonberger2016structure, mildenhall2021nerf, kerbl3Dgaussians, wang2025vggt}. Despite these advances, relatively few works directly operate on TSDF based shape completions, even though TSDFs can be easily obtained from RGB-D data through standard fusion pipelines. Early learning-based approaches such as 3D-EPN~\citep{3depn} employ 3D encoder--decoder architectures to directly predict complete shapes from partial volumetric inputs. Subsequent methods, including Few-Shot completion~\citep{fewshotcompletion} and IF-Nets~\citep{ifnets}, improve generalization by using learned priors and implicit representations, while Auto-SDF~\citep{autosdf} introduces generative modeling of signed distance functions to capture shape distributions. Generative methods such as SDF-StyleGAN~\citep{zheng2022sdf} models signed distance functions using a generative adversarial framework to complete 3D shapes, while RePaint-3D~\citep{lugmayr2022repaint} formulates 3D shape completion as a diffusion-based inpainting task. 

As deterministic approach PatchComplete~\citep{tang2022patchcomplete} learns multi-resolution local patch priors for TSDF-based completion and demonstrates strong generalization to unseen categories, but relies on multi-stage training and limited global reasoning. Recent stochastic models~\citep{chu2023diffcomplete, schaefer2024sc} formulate shape completion as a diffusion process, enabling globally consistent outputs at the cost of iterative sampling and higher computational complexity. BridgeShape~\citep{kong2026bridgeshape} performs completion through a latent diffusion Schrödinger bridge\citep{liu2023i2sb}, using DINOv2\citep{oquab2023dinov2} features to enrich a pretrained VQ-VAE\citep{razavi2019generating}. However, its semantic priors remain implicit in the latent representation. Another recent work ~\citep{zakeri2026latent} introduce uncertainty-aware SDF latent transformers trained on large-scale multi-view data~\citep{deitke2023objaverse} to improve completion quality but they still lack global context between shape categories. To address the need for efficient long-range modeling in 3D, state space models (SSMs) have recently emerged as a promising alternative, with Voxel Mamba~\citep{zhang2024voxel} adapting Mamba-style sequence modeling to voxelized representations. Building on these advances, our approach focuses on TSDF-based completion by combining deterministic reconstruction with voxelized semantic priors and efficient state space modeling, enabling consistent completion while maintaining low computational cost.

\section{Method}
Our goal is to reconstruct a complete 3D shape from an incomplete TSDF scan. Given a partial scan $X_{partial} \in \mathbb{R}^{1 \times D \times H \times W}$, we predict the complete scan $X_{pred} \in \mathbb{R}^{1 \times D \times H \times W}$ by combining geometric features with learned semantic priors. While TSDF representations provide local surface information, they lack global understanding due to occlusions. To address this, we introduce TSDF-DINO features that capture global object structure. We train a \textbf{TSDF-only} 3D student to predict voxel-aligned DINO features directly from incomplete geometry, using supervision constructed from multi-view images with a frozen, pretrained DINOv3 teacher. We then integrate these distilled features into our Multi-Scale Voxel Mamba (MSM) module for efficient global context aggregation in shape completion. At inference, the model requires only the partial TSDF, without RGB images or DINOv3 execution. Preliminaries for our TSDF-DINO model and SSMs are provided in the supplementary material.

\begin{figure}[t]
    \centering
    \includegraphics[width=\linewidth]{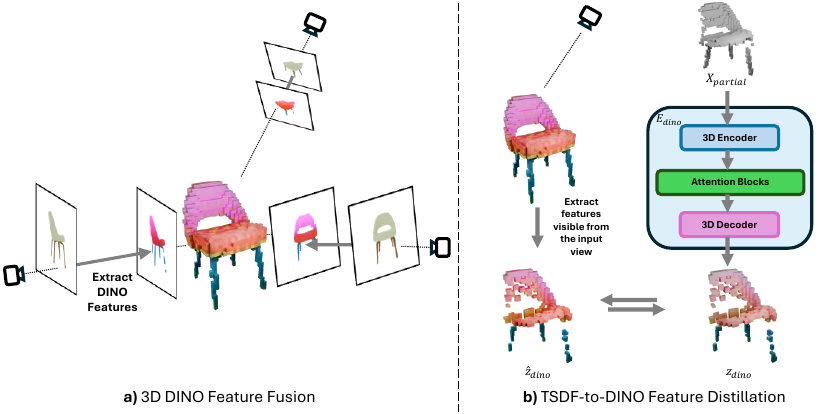}
    \caption{
Overview of the TSDF-DINO distillation pipeline. 
(a) DINO features extracted from multi-view images are fused into a voxel-aligned 3D feature grid. 
(b) A student network is trained to predict semantic features directly from partial TSDF inputs via distillation.
}
\label{fig:dino_preparation}
\end{figure}

\subsection{TSDF-DINO Distillation}

We train a 3D student network that predicts voxel-wise semantic features from partial TSDF inputs. Given an input TSDF $X_{partial}$ the student encoder predicts a feature volume with channel size $C$:
\begin{equation}
z_{\text{dino}} = E_{\text{dino}}(X_{partial}), \quad z_{\text{dino}} \in \mathbb{R}^{C \times D \times H \times W}.
\end{equation}


To supervise this model, we use an externally pretrained DINOv3~\cite{simeoni2025dinov3} teacher. For each ShapeNet training shape, we render 20 RGB-D views, extract dense image features and fuse them into a semantic volume aligned with the TSDF grid. For each partial TSDF input, we apply the coverage mask of its corresponding input view to this fused volume, obtaining the partial teacher target \(\hat z_{\mathrm{dino}}\). The student therefore learns to predict semantic features aligned with the observed geometry.


The student predicts a gating mask \(m \in [0,1]^{1\times D\times H\times W}\) to suppress unreliable output features. We use \(\mathcal L_{\mathrm{mask}}=\mathrm{BCE}(m,\hat m)\), where the binary target \(\hat m\) marks valid nonzero teacher features and is used only during training. See Appendix A.1.1 for teacher-volume construction and masking.

We train the model using cosine similarity, reconstruction, and mask supervision losses:
\begin{equation}
\mathcal{L}_{\text{cos}} = 1 - \frac{\langle z_{\text{dino}}, \hat{z}_{\text{dino}} \rangle}{\|z_{\text{dino}}\|_2 \cdot \|\hat{z}_{\text{dino}}\|_2}, \quad
\mathcal{L}_{\text{mse}} = \| z_{\text{dino}} - \hat{z}_{\text{dino}} \|_2^2, \quad
\mathcal{L}_{\text{mask}} = \mathrm{BCE}(m, \hat{m}),
\end{equation}
\begin{equation}
\mathcal{L}_{\text{distill}} = 
\lambda_{\text{cos}} \mathcal{L}_{\text{cos}} +
\lambda_{\text{mse}} \mathcal{L}_{\text{mse}} +
\lambda_{\text{mask}} \mathcal{L}_{\text{mask}}.
\end{equation}
The cosine loss $\mathcal{L}_{\text{cos}}$ aligns the semantic direction of features, the reconstruction loss $\mathcal{L}_{\text{mse}}$ enforces voxel-wise magnitude alignment, and the binary cross-entropy loss $\mathcal{L}_{\text{mask}}$ supervises the gating mask to identify valid semantic regions and suppresses noise. This distillation process transfers semantic priors from 2D DINO representations into a geometrically aligned 3D feature space. Trained on partial TSDF observations, the distilled student provides global context for shape understanding without requiring RGB inputs or DINOv3 execution at test time, enabling TSDF-only inference.

\begin{figure}[t]
    \centering
    \includegraphics[width=\linewidth]{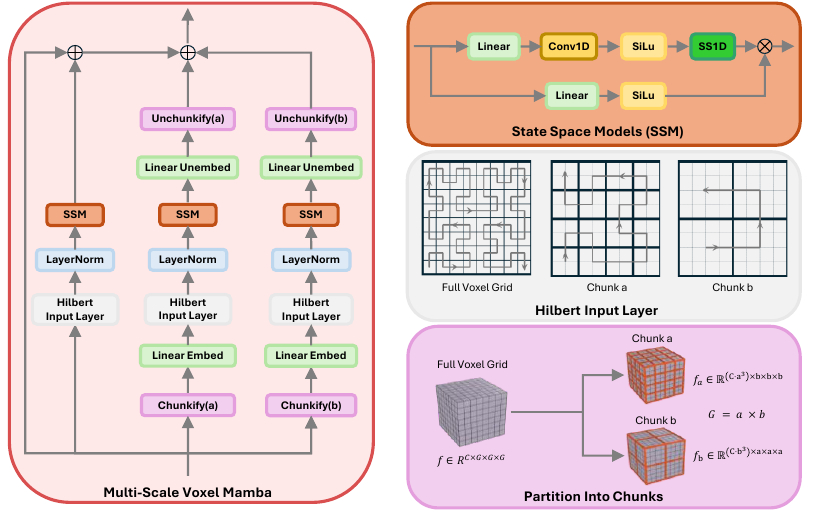}
    \caption{
Overview of our Multi-Scale Voxel Mamba architecture. Starting from a full voxel grid, the input is decomposed into multiple chunks using chunk sizes $a$ and $b$. Each chunk is embedded and processed independently through Hilbert-ordered input layers followed by SSM blocks. The processed features are then projected back into the original voxel grid, with a residual connection applied across the branches.}
\label{fig:MSM32}
\end{figure}

\subsection{Voxel State Space Modeling for Fusion and Multi-Scale Refinement}
To model long-range dependencies in dense TSDF feature volumes, we adopt a voxel state space formulation based on Voxel Mamba~\cite{zhang2024voxel}, with multi-scale patch reasoning inspired by PatchComplete~\cite{tang2022patchcomplete}. We serialize voxel features using a Hilbert curve and apply Mamba-based SSMs~\cite{gu2023mamba} for efficient sequence modeling with linear complexity in sequence length. Voxel Mamba processes sparse outdoor point clouds using state space modeling at reduced spatial resolution. Our full-grid refinement instead processes the dense $32^3$ feature grid directly, obtaining long-range context through state propagation without additional downsampling. We define a general voxel state space operator:
\begin{equation}
\phi(f) = \big(\mathrm{SSM}(\mathrm{LN}(\mathrm{HIL}(f)))\big)
\end{equation}
where \(f \in \mathbb{R}^{C \times G \times G \times G}\) denotes voxel features, 
\(\mathrm{HIL}(\cdot)\) maps the 3D grid into a 1D sequence using a Hilbert curve~\cite{hilbert1935stetige} to preserve spatial locality, 
\(\mathrm{LN}(\cdot)\) denotes Layer Normalization~\cite{ba2016layer}, 
and \(\mathrm{SSM}(\cdot)\) denotes a Mamba-based state space model~\cite{gu2023mamba}. 
The output is reshaped back to the voxel grid, enabling global context aggregation while preserving fine-grained geometric details.

\paragraph{Multi-Scale Voxel Mamba.}
\label{sec:refinement}

To enhance spatial reasoning at high resolution, we propose a multi-scale extension of the voxel state space operator. As illustrated in Figure~\ref{fig:MSM32}, full-grid sequence modeling using $\phi(\cdot)$ is combined with parallel chunk-wise branches that represent local voxel regions as tokens. These branches model interactions across the chunk grid using shorter sequences, providing regional context complementary to voxel-level processing. Their outputs are restored to the original voxel layout before aggregation.

Given a feature volume $f \in \mathbb{R}^{C \times G \times G \times G}$, we partition it into non-overlapping 3D chunks of size $R \times R \times R$, where $R \in \{a,b\}$ and $G = a \cdot b$. This yields two chunkified representations:
\begin{equation}
f_a \in \mathbb{R}^{(C \cdot a^3) \times b \times b \times b}, 
\quad
f_b \in \mathbb{R}^{(C \cdot b^3) \times a \times a \times a}
\end{equation}
where each spatial location corresponds to a flattened local neighborhood.

Each chunk is embedded into a token representation, serialized using a Hilbert curve over the chunk grid, and processed with the state space operator. The chunk-level operator is defined as:
\begin{equation}
\psi_R(f) = \mathrm{Unchunkify_R}\big(\mathrm{Unembed}(\phi(\mathrm{Embed}(f_R)))\big),
\end{equation}
where $\mathrm{Embed}(\cdot)$ maps each flattened chunk of dimension into a lower-dimensional token space via a linear projection,  
$\mathrm{Unembed}(\cdot)$ projects tokens back to the original chunk feature space, and  
$\mathrm{Unchunkify_R}(\cdot)$ restores the original spatial layout by rearranging chunks to the $G \times G \times G$ grid. The chunked branches operate on reduced-length sequences of $a^3$ and $b^3$ tokens, enabling efficient global reasoning across spatially distant regions. In contrast, the full-resolution branch $\phi(f)$ processes the complete voxel sequence, retaining original spatial structure.
Finally, we fuse multi-scale features using residual aggregation:
\begin{equation}
\lambda(f) = \phi(f) + \psi_a(f) + \psi_b(f) + f
\end{equation}

\subsection{Shape Completion Architecture}
\paragraph{Feature encoding.}
Given an incomplete TSDF input $X_{\text{partial}}$, we extract two complementary voxel-aligned feature representations:
\begin{equation}
z_{\text{tsdf}} = E_{\text{tsdf}}(X_{\text{partial}}), \qquad
z_{\text{dino}} = E_{\text{dino}}(X_{\text{partial}}),
\end{equation}
where $E_{\text{tsdf}}(\cdot)$ learns geometric features directly from the TSDF input, and $E_{\text{dino}}(\cdot)$ is initialized from the pretrained DINO-TSDF  distillation model that produces voxel-aligned semantic features. We fine-tune $E_{\text{dino}}$ within the shape completion network using the same shape completion objective.
\begin{figure}[t]
    \centering
    \includegraphics[width=\linewidth]{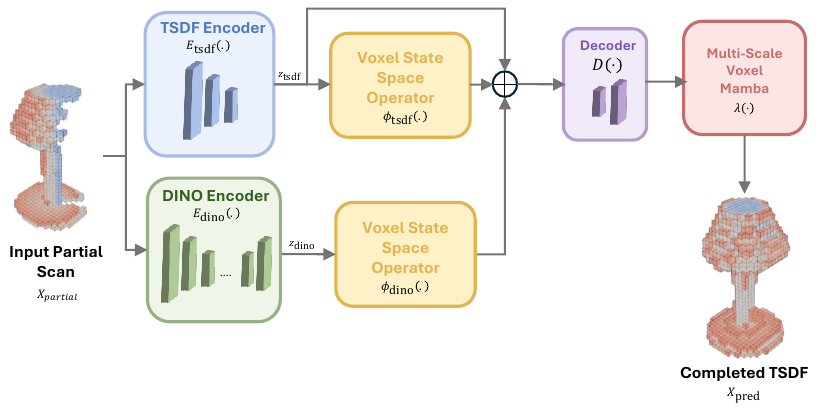}
    \caption{Overview of the shape completion pipeline.}
    \label{fig:architecture}
\end{figure}
\paragraph{Cross-modal fusion.}
At an intermediate decoder resolution, we are given a geometric feature volume \(z_{\text{tsdf}}\) and an aligned semantic feature volume \(z_{\text{dino}}\). We apply the voxel state space operator independently to each feature stream before fusion:
\begin{equation}
z_{\text{fused}} = \phi_\text{tsdf}(z_{\text{tsdf}}) + \phi_\text{dino}(z_{\text{dino}}) + z_{\text{tsdf}}.
\end{equation}
While both feature volumes are derived from partial observations, \(z_{\text{dino}}\) encodes filtered semantic cues that largely reflect only the observed regions, whereas \(z_{\text{tsdf}}\) retains the full geometric signal, including noisy and uncertain areas. The residual connection \(z_{\text{tsdf}}\) ensures that the fused representation remains grounded in the underlying TSDF geometry, preventing over-reliance on the semantic branch.

\paragraph{Decoding and refinement.}
The fused representation is decoded using a 3D CNN-based decoder  $D(\cdot)$ that  recovers spatial resolution and propagates feature representations. To further propagate information across distant regions, we apply the proposed MSM module:
\begin{equation}
X_{\text{pred}} = \lambda(D(z_{fused})),
\end{equation}
where $\lambda(\cdot)$ aggregates full-resolution and chunk-wise sequence modeling explained in \ref{sec:refinement}. The final prediction $X_{\text{pred}} \in \mathbb{R}^{32 \times 32 \times 32}$ represents the completed TSDF volume. The overall architecture of our shape completion model is shown in Figure \ref{fig:architecture}.

\paragraph{Loss.}
We use the sign-aware weighted TSDF reconstruction loss used in ~\cite{tang2022patchcomplete}. We first define occupancy masks based on the sign of the TSDF and then construct error masks as follows:
\begin{equation}
\begin{gathered}
M_{\text{gt}} = (X_{\text{gt}} \leq 0), \qquad
M_{\text{pred}} = (X_{\text{pred}} \leq 0), \\
M_{\text{fp}} = M_{\text{pred}} (1 - M_{\text{gt}}), \qquad
M_{\text{fn}} = (1 - M_{\text{pred}}) M_{\text{gt}}, \qquad
M_{\text{correct}} = 1 - M_{\text{fp}} - M_{\text{fn}}.
\end{gathered}
\end{equation}

Using these masks, we define the loss function as
\begin{equation}
\mathcal{L}_{\text{tsdf}} =
w_{\text{fn}} M_{\text{fn}} \ell(X_{\text{pred}}, X_{\text{gt}})
+
w_{\text{fp}} M_{\text{fp}} \ell(X_{\text{pred}}, X_{\text{gt}})
+
w_{\text{correct}} M_{\text{correct}} \ell(X_{\text{pred}}, X_{\text{gt}}),
\end{equation}
where $\ell$ is the voxel-wise Smooth-$\ell_1$ loss. We use $w_{\text{fn}}=5$, $w_{\text{fp}}=3$, and $w_{\text{correct}}=1$ following ~\citep{tang2022patchcomplete}.

\section{Experiments}

We focus our evaluation on unseen categories to highlight the strength of our method in using DINO features for capturing global semantics and generalizing beyond the training distribution. For qualitative comparisons, we include PatchComplete~\cite{tang2022patchcomplete} and DiffComplete~\cite{chu2023diffcomplete} as representative baselines, as both provide publicly available implementations and are widely adopted in TSDF-based shape completion. We additionally include the published quantitative results of SC-Diff~\cite{schaefer2024sc} and BridgeShape~\cite{kong2026bridgeshape}, as complete implementations and pretrained checkpoints were not publicly available when we conducted our experiments. Additional experiments, including results on seen categories~\citep{3depn} and limitations of our model, are provided in the supplementary material for completeness.

\subsection{Setup}
\paragraph{Dataset.} We train and evaluate our approach on synthetic shape data from ShapeNet~\cite{chang2015shapenet} and real-world scan data from ScanNet~\cite{dai2017scannet}. For ShapeNet, we follow the data generation pipeline of prior work~\citep{tang2022patchcomplete} and use virtually scanned partial observations as input, with corresponding complete shapes as ground truth. For ScanNet, we use real scanned objects extracted via bounding boxes, with complete targets provided by Scan2CAD~\cite{avetisyan2019scan2cad} alignment. For training the TSDF-DINO student model, we obtain ground-truth 3D semantic features by fusing multi-view 2D DINO ~\cite{simeoni2025dinov3} features into voxel grids aligned with the corresponding shapes. We follow the same train/test split used in ~\citep{tang2022patchcomplete, chu2023diffcomplete} for our feature distillation model and shape completion model. For all experiments, objects are represented as $32^3$ TSDF volumes with a truncation value of 3 voxel units. 

\paragraph{Training.}
We first train our DINO student model on ShapeNet for 30 epochs using 5 A40 GPUs, a batch size of 4, and an initial learning rate of $10^{-4}$ with the Adam optimizer, which takes approximately 2 hours. The learned weights are then used to initialize the DINO branch in our shape completion model. 

We train our shape completion model on ShapeNet for 80 epochs under the same training setup which takes approximately 2 hours. Finally, we fine-tune the model on ScanNet for 2 epochs, where only the TSDF-DINO branch and the encoder of the TSDF branch are updated, while the remaining components are frozen. Fine-tuning takes approximately 3 minutes.

\subsection{Results}

\begin{figure}[t]
    \centering
    \includegraphics[width=0.9\linewidth]{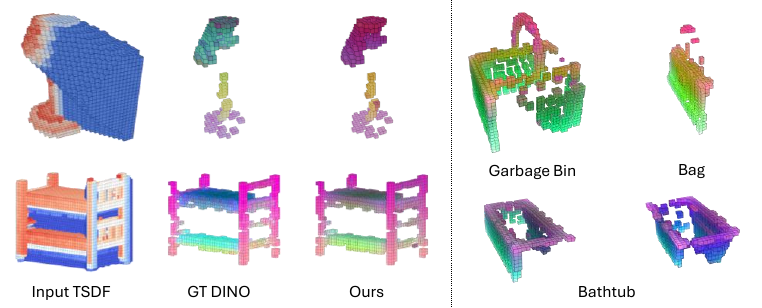}
    \caption{TSDF-DINO model qualitative results on unseen objects.}
    \label{fig:shapenet-dino-results}
\end{figure}

\paragraph{Distillation Results.} We present qualitative results of our TSDF-DINO model in Figure~\ref{fig:shapenet-dino-results}. Instead of rendering incomplete shapes, we visualize TSDF values in the range $[-\textit{truncation}, +\textit{truncation})$ to better expose noise patterns. Red indicates values near $+\text{truncation}$, while blue indicates values near $-\textit{truncation}$, with dark blue representing occluded regions. We exclude $+\textit{truncation}$ values to avoid clutter from empty regions, while retaining $-\textit{truncation}$ values to capture interior and occlusion. The resulting spatial stretching indicates the incompleteness of single-view TSDF generation.

For feature visualization, we project the features into a 3D color space using Principal Component Analysis (PCA)~\cite{abdi2010principal}. On the left, we compute a shared PCA embedding for the ground-truth DINO features and their corresponding predicted features to enable direct comparison, showing that our model produces similar results. On the right,  we compute one PCA for semantically related but different categories(garbage bin and bag) and another PCA for objects within the same category(bathtubs). The resulting feature projections show that semantically similar parts exhibit similar feature patterns. Importantly, all results are obtained on categories not used during distillation training, indicating that the TSDF-DINO model captures global semantic structures.


\paragraph{Quantitative Results.} As shown in Table~\ref{tab:synthetic_unseen}, our method achieves the best average performance across unseen synthetic ShapeNet~\cite{chang2015shapenet} categories in both Chamfer Distance (CD) and Intersection over Union (IoU). In particular, we consistently improve over both PatchComplete~\cite{tang2022patchcomplete} and DiffComplete~\cite{chu2023diffcomplete} across most categories, while also outperforming SC-Diff~\cite{schaefer2024sc} and BridgeShape~\cite{kong2026bridgeshape} on average. In Table~\ref{tab:real_unseen}, we further show that on real-world ScanNet~\cite{dai2017scannet} data, our method achieves competitive average performance. Despite being trained primarily on synthetic data, our model adapts to real scans with brief fine-tuning, demonstrating strong generalization across different noise characteristics. In Table~\ref{tab:efficiency}, we report the number of model parameters, GPU memory usage during inference, and inference time for one sample. Our method achieves lower memory usage and faster inference than the compared methods while maintaining competitive accuracy across unseen categories. DiffComplete requires costly iterative denoising, while deterministic approaches like PatchComplete rely on heavy local priors. Our model remains lightweight and efficient while achieving strong completion performance. This is enabled by our distilled semantic DINO representations, which provide strong global priors, together with our multi-scale refinement that captures long-range information at low computational cost.

\begin{table}[t]
\centering
\caption{Shape completion results on synthetic objects of unseen categories. ·/· are CD($\downarrow$) / IoU($\uparrow$).}
\label{tab:synthetic_unseen}
\normalsize
\setlength{\tabcolsep}{3pt}
\begin{adjustbox}{max width=\linewidth}
\begin{tabular}{lccccccccc}
\toprule
Category
& \makecell{3D-\\EPN~\citep{3depn}}
& \makecell{Few-\\Shot~\citep{fewshotcompletion}}
& \makecell{IF-\\Nets~\citep{ifnets}}
& \makecell{Auto-\\SDF~\citep{autosdf}}
& \makecell{Patch-\\Complete~\citep{tang2022patchcomplete}}
& \makecell{Diff-\\Complete~\citep{chu2023diffcomplete}}
& \makecell{SC-\\Diff~\citep{schaefer2024sc}}
& \makecell{Bridge-\\Shape~\citep{kong2026bridgeshape}}
& Ours \\
\midrule
Bag
& 5.01 / 73.8 & 8.00 / 56.1 & 4.77 / 69.8 & 5.81 / 56.3
& 3.94 / 77.6 & 3.86 / 78.3 & 3.79 / 78.3
& \textbf{3.70} / \textbf{80.0} & 3.84 / 77.9 \\

Lamp
& 8.07 / 47.2 & 15.1 / 25.4 & 5.70 / 50.8 & 6.57 / 39.1
& 4.68 / 56.4 & 4.80 / 57.9 & 4.74 / 60.0
& 4.94 / 62.7 & \textbf{3.78} / \textbf{64.3} \\

Bathtub
& 4.21 / 57.9 & 7.05 / 45.7 & 4.72 / 55.0 & 5.17 / 41.0
& 3.78 / 66.3 & 3.52 / 68.9 & 3.67 / 65.9
& 3.41 / 69.8 & \textbf{3.21} / \textbf{73.3} \\

Bed
& 5.84 / 58.4 & 10.0 / 39.6 & 5.34 / 60.7 & 6.01 / 44.6
& 4.49 / 66.8 & 4.16 / 67.1 & 4.40 / 67.1
& 4.32 / 69.6 & \textbf{3.92} / \textbf{71.6} \\

Basket
& 7.90 / 54.0 & 8.72 / 40.6 & 4.44 / 50.2 & 6.70 / 39.8
& 5.15 / 61.0 & 4.94 / 65.5 & 4.89 / \textbf{68.5}
& 5.50 / 65.1 & \textbf{3.96} / 67.9 \\

Printer
& 5.15 / 73.6 & 9.26 / 56.7 & 5.83 / 70.5 & 7.52 / 49.9
& 4.63 / 77.6 & 4.40 / 76.8 & 4.36 / 76.8
& \textbf{4.09} / \textbf{81.4} & 4.17 / 80.2 \\

Laptop
& 3.90 / 62.0 & 10.4 / 31.3 & 6.47 / 58.3 & 4.81 / 51.1
& 3.77 / 63.8 & 3.52 / 67.4 & 3.41 / 68.4
& \textbf{3.13} / \textbf{72.7} & 3.21 / 71.6 \\

Bench
& 4.54 / 48.3 & 8.11 / 27.2 & 5.03 / 49.7 & 4.31 / 39.5
& 3.70 / 53.9 & 3.56 / 58.2 & 3.39 / 61.1
& 3.38 / 59.1 & \textbf{3.02} / \textbf{62.1} \\
\midrule
Average
& 5.58 / 59.4 & 9.58 / 40.3 & 5.29 / 58.1 & 5.86 / 45.2
& 4.27 / 65.4 & 4.10 / 67.5 & 4.08 / 68.2
& 4.06 / 70.1 & \textbf{3.64} / \textbf{71.1} \\

& {$\pm 2\mathrm{e}^{-1}$/$\pm 8\mathrm{e}^{-1}$}
& {$\pm 1\mathrm{e}^{-1}$/$\pm 1\mathrm{e}^{-1}$}
& {$\pm 1\mathrm{e}^{-1}$/$\pm 3\mathrm{e}^{-1}$}
& {$\pm 5\mathrm{e}^{-3}$/$\pm 7\mathrm{e}^{-1}$}
& {$\pm 5\mathrm{e}^{-2}$/$\pm 1\mathrm{e}^{-1}$}
& {$\pm 2\mathrm{e}^{-2}$/$\pm 3\mathrm{e}^{-1}$}
& --
& --
& {$\pm 3\mathrm{e}^{-2}$/$\pm 1\mathrm{e}^{-1}$} \\
\bottomrule
\end{tabular}
\end{adjustbox}
\end{table}

\begin{table}[t]
\centering
\caption{Shape completion results on real-world objects of unseen categories. ·/· are CD($\downarrow$) / IoU($\uparrow$).}
\label{tab:real_unseen}
\normalsize
\setlength{\tabcolsep}{2pt}
\begin{adjustbox}{max width=\linewidth}
\begin{tabular}{lccccccccc}
\toprule
Category
& \makecell{3D-\\EPN~\citep{3depn}}
& \makecell{Few-\\Shot~\citep{fewshotcompletion}}
& \makecell{IF-\\Nets~\citep{ifnets}}
& \makecell{Auto-\\SDF~\citep{autosdf}}
& \makecell{Patch-\\Complete~\citep{tang2022patchcomplete}}
& \makecell{Diff-\\Complete~\citep{chu2023diffcomplete}}
& \makecell{SC-\\Diff~\citep{schaefer2024sc}}
& \makecell{Bridge-\\Shape~\citep{kong2026bridgeshape}}
& Ours \\
\midrule
Bag
& 8.83 / 53.7 & 9.10 / 44.9 & 8.96 / 44.2 & 9.30 / 48.7
& 8.23 / 58.3 & 7.05 / 48.5 & 7.41 / 50.0
& 7.67 / \textbf{61.2} & \textbf{7.00} / 60.9 \\

Lamp
& 14.3 / 20.7 & 11.9 / 19.6 & 10.2 / 24.9 & 11.2 / 24.4
& 9.42 / 28.4 & 6.84 / 30.5 & \textbf{6.39} / 33.2
& 8.07 / 36.3 & 8.33 / \textbf{37.8} \\

Bathtub
& 7.56 / 41.0 & 7.77 / 38.2 & 7.19 / 39.5 & 7.84 / 36.6
& 6.77 / 48.0 & 8.22 / 48.5 & 8.09 / 48.5
& \textbf{6.28} / \textbf{50.9} & 6.31 / 50.6 \\

Bed
& 7.76 / 47.8 & 9.07 / 34.9 & 8.24 / 44.9 & 7.91 / 38.0
& 7.24 / 48.4 & 7.20 / 46.6 & 6.91 / 48.6
& \textbf{6.87} / \textbf{51.3} & 6.99 / 49.5 \\

Basket
& 7.74 / 36.5 & 8.02 / 34.3 & 6.74 / 42.7 & 7.54 / 36.1
& 6.60 / 45.5 & 7.42 / 59.2 & 6.38 / \textbf{62.2}
& \textbf{6.20} / 50.2 & 6.35 / 46.5 \\

Printer
& 8.36 / 63.0 & 8.30 / 62.2 & 8.28 / 60.7 & 9.66 / 49.9
& 6.84 / 70.5 & \textbf{6.36} / \textbf{74.5} & 7.10 / 69.1
& 6.83 / 71.0 & 6.59 / 72.5 \\
\midrule
Avg.
& 9.09 / 44.0 & 9.02 / 38.6 & 8.26 / 42.6 & 8.90 / 38.9
& 7.52 / 49.5 & 7.18 / 51.3 & 7.04 / 51.9
& \textbf{6.99} / \textbf{53.5} & 7.09 / 53.0 \\

& {$\pm 3\mathrm{e}^{-1}$/$\pm 3\mathrm{e}^{-1}$}
& {$\pm 8\mathrm{e}^{-2}$/$\pm 6\mathrm{e}^{-1}$}
& {$\pm 8\mathrm{e}^{-2}$/$\pm 7\mathrm{e}^{-1}$}
& {$\pm 2\mathrm{e}^{-2}$/$\pm 3\mathrm{e}^{-1}$}
& {$\pm 2\mathrm{e}^{-2}$/$\pm 5\mathrm{e}^{-1}$}
& {$\pm 4\mathrm{e}^{-2}$/$\pm 3\mathrm{e}^{-1}$}
& --
& --
& {$\pm 2\mathrm{e}^{-2}$/$\pm 2\mathrm{e}^{-1}$} \\
\bottomrule
\end{tabular}
\end{adjustbox}
\end{table}

\begin{figure}[t]
    \centering
    \includegraphics[width=0.97\linewidth]{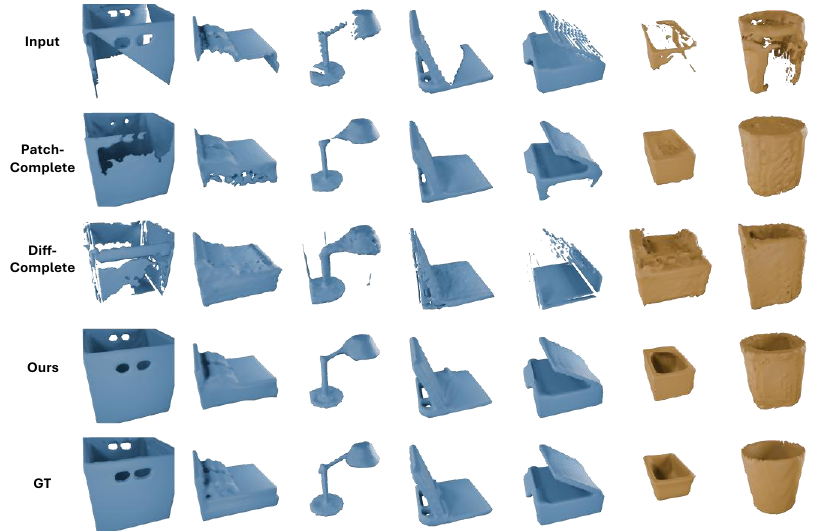}
    \caption{Shape completion results on both synthetic (blue) and real-world (yellow) objects from entirely unseen categories. Our method generates high-quality completed shapes across both datasets.}
    \label{fig:results}
\end{figure}

\paragraph{Qualitative Results.} Figure~\ref{fig:results} presents qualitative comparisons on a set of unseen objects, including both synthetic~\cite{chang2015shapenet} and real-world~\cite{dai2017scannet} shapes. We observe that existing methods such as PatchComplete and DiffComplete tend to either overfit to the visible partial input or lose important structural details in the observed regions, such as the holes in the basket and the back side of the laptop. In contrast, our method better preserves these observed regions while completing the missing parts. Previous methods can also produce noisy reconstructions in more challenging real-world cases, as seen in the real world examples, where artifacts become more apparent. Overall, our method preserves fine geometric details while maintaining clean and coherent completions, achieving a better balance between fitting the observed input and reconstructing missing regions.

\subsection{Ablation Studies}
All ablations in this section use the ShapeNet\citep{chang2015shapenet} test split for unseen categories and follow the evaluation protocol of Table~\ref{tab:synthetic_unseen}.

\paragraph{Feature Modules} In Table~\ref{tab:module_ablation} we evaluate the contribution of our proposed modules. The sole TSDF row follows a standard U-Net~\citep{ronneberger2015u} architecture and we select this as baseline. The DINO branch row, is our dino model and a prediction head on top of pretrained features. We see that using the DINO branch alone leads to overfitting to the incomplete input geometry, limiting its ability to infer missing regions. Combining TSDF and DINO branches improves results by combining geometric features with semantic reasoning. Finally, incorporating the MSM module further enhances results by enabling effective global context aggregation. 

\paragraph{Tuning DINO module.}
In Table~\ref{tab:dino_strategy}, we compare three training strategies for the DINO module in our shape completion architecture. Initializing the DINO encoder with random weights leads to significantly degraded performance, indicating that the gains do not stem from increased model capacity, but rather from the strong semantic priors encoded in the pretrained DINO features. Using frozen pretrained weights yields competitive results, demonstrating that these priors already provide useful guidance for shape completion. However, this also supports our earlier observation that relying solely on the DINO branch may lead to overfitting to incomplete geometry. Finetuning further improves performance, highlighting the importance of adapting DINO features to completion task.

\paragraph{Effect of Training Data Ratio.}
In Table~\ref{tab:train_split_ablation}, we study the impact of varying the amount of training data by using different percentages of the standard training dataset. DiffComplete~\cite{chu2023diffcomplete} degrades significantly in low-data ratio, indicating its reliance on large-scale supervision to learn stable shape distributions. In contrast, our method remains robust across all settings, maintaining strong performance even with reduced data and achieving SOTA results even at intermediate ratios. We attribute this data efficiency to voxelized DINO features, which provide strong semantic and structural priors for consistent generalization.

\paragraph{Feature connection between branches.}We compare different strategies for fusing TSDF and semantic DINO features in Table~\ref{tab:dino_connection_ablation}. Simple concatenation followed by convolution provides a basic fusion mechanism, while attention-based fusion increases memory usage without significant performance gains. In contrast, our voxel state residual fusion achieves the best results, effectively balancing geometric and semantic information.

\newsavebox{\efficiencytablebox}
\newsavebox{\componentstablebox}
\newsavebox{\dinostrategytablebox}

\begin{table*}[t]
\centering
\small
\setlength{\tabcolsep}{3.5pt}
\renewcommand{\arraystretch}{1.05}


\begin{lrbox}{\efficiencytablebox}
\begin{minipage}[t]{0.48\textwidth}
\centering
\captionof{table}{Model efficiency comparison.}
\vspace{-6pt}
\label{tab:efficiency}
\begin{tabular}{lccc}
\toprule
Method & Params & Mem. & Time \\
 & (M) & (GB) & (ms) \\
\midrule
PatchComplete~\citep{tang2022patchcomplete}
& 182.85 & 1.259 & 65.56 \\
DiffComplete~\citep{chu2023diffcomplete}
& 43.09 & 0.302 & 3115.85 \\
Ours
& \textbf{25.97} & \textbf{0.292} & \textbf{28.68} \\
\bottomrule
\end{tabular}
\end{minipage}
\end{lrbox}

\begin{lrbox}{\componentstablebox}
\begin{minipage}[t]{0.48\textwidth}
\centering
\captionof{table}{Ablation study of the components.}
\vspace{-6pt}
\label{tab:module_ablation}
\begin{tabular}{c c c | c c}
\toprule
TSDF & DINO & MSM & CD $\downarrow$ & IoU $\uparrow$ \\
\midrule
\checkmark &  &  & 4.29 & 66.1 \\
 & \checkmark &  & 3.79 & 70.1 \\
\checkmark & \checkmark &  & 3.70 & 70.2 \\
\checkmark &  & \checkmark & 3.98 & 67.6 \\
 & \checkmark & \checkmark & 3.76 & 70.4 \\
\checkmark & \checkmark & \checkmark
& \textbf{3.64} & \textbf{71.1} \\
\bottomrule
\end{tabular}
\end{minipage}
\end{lrbox}

\begin{lrbox}{\dinostrategytablebox}
\begin{minipage}[t]{0.48\textwidth}
\centering
\captionof{table}{Impact of different training strategies.}
\label{tab:dino_strategy}
\begin{tabular}{l | c c}
\toprule
DINO Strategy & CD $\downarrow$ & IoU $\uparrow$ \\
\midrule
Random Init. & 4.16 & 65.8 \\
Frozen & 4.05 & 67.7 \\
Finetuned & \textbf{3.64} & \textbf{71.1} \\
\bottomrule
\end{tabular}
\end{minipage}
\end{lrbox}

\begin{minipage}[t]{0.48\textwidth}
\centering
\vspace*{25pt}

\usebox{\componentstablebox}
\par

\captionof{table}{Shape results on unseen categories using different training dataset ratio.}
\label{tab:train_split_ablation}
\resizebox{\linewidth}{!}{%
\begin{tabular}{lccc}
\toprule
Method & 100\% Train & 50\% Train & 20\% Train \\
\midrule
PatchComplete~\citep{tang2022patchcomplete}
& 4.27 / 65.4 & 4.43 / 63.5 & 5.10 / 61.2 \\
DiffComplete~\citep{chu2023diffcomplete}
& 4.10 / 67.5 & 9.23 / 43.8 & 19.46 / 11.8 \\
Ours
& \textbf{3.64} / \textbf{71.1}
& \textbf{3.97} / \textbf{67.8}
& \textbf{4.35} / \textbf{64.4} \\
\bottomrule
\end{tabular}%
}

\end{minipage}
\hfill
\begin{minipage}[t]{0.48\textwidth}
\centering
\vspace*{6pt}

\usebox{\efficiencytablebox}
\par

\usebox{\dinostrategytablebox}
\par

\captionof{table}{Ablation on feature connection.}
\label{tab:dino_connection_ablation}
\begin{tabular}{lc}
\toprule
Method & CD / IoU \\
\midrule
Concat Conv & 3.87 / 69.6 \\
Attention & 3.86 / 69.2 \\
Voxel State Residual (Ours) & \textbf{3.64} / \textbf{71.1} \\
\bottomrule
\end{tabular}

\end{minipage}
\end{table*}

\section{Conclusion}
We introduced DinoComplete, the first shape completion framework that uses TSDF-guided distilled DINO semantic priors and voxel state space models. By predicting voxel-aligned semantic features directly from partial TSDFs, our method reconstructs complete 3D shapes from partial scans without requiring additional inputs at inference time. DinoComplete achieves strong shape completion performance while maintaining an efficient and deterministic pipeline. We believe that integrating distilled semantic priors with structured 3D representations is an important step toward more robust and generalizable real-world shape completion and understanding, and we hope this work inspires further research on unconventional representations.

\section{Acknowledgements and Disclosures}
This work was supported by the German Federal Ministry of Research, Technology and Space (BMFTR) through the Robotics Institute Germany (RIG) and by the Bavarian Ministry of Economic Affairs, Regional Development and Energy as part of the project “6G Future Lab Bavaria”.

\bibliographystyle{plain} 
\bibliography{references}

@inproceedings{newcombe2011kinectfusion,
  title={Kinectfusion: real-time 3d reconstruction and interaction using a moving depth camera},
  author={Izadi, Shahram and Kim, David and Hilliges, Otmar and Molyneaux, David and Newcombe, Richard and Kohli, Pushmeet and Shotton, Jamie and Hodges, Steve and Freeman, Dustin and Davison, Andrew and others},
  booktitle={Proceedings of the 24th annual ACM symposium on User interface software and technology},
  pages={559--568},
  year={2011}
}

@inproceedings{dai2017scannet,
    title={ScanNet: Richly-annotated 3D Reconstructions of Indoor Scenes},
    author={Dai, Angela and Chang, Angel X. and Savva, Manolis and Halber, Maciej and Funkhouser, Thomas and Nie{\ss}ner, Matthias},
    booktitle = {Proc. Computer Vision and Pattern Recognition (CVPR), IEEE},
    year = {2017}
}

@inproceedings{song2015sun,
  title={Sun rgb-d: A rgb-d scene understanding benchmark suite},
  author={Song, Shuran and Lichtenberg, Samuel P and Xiao, Jianxiong},
  booktitle={Proceedings of the IEEE conference on computer vision and pattern recognition},
  pages={567--576},
  year={2015}
}

@article{chang2017matterport3d,
  title={Matterport3D: Learning from RGB-D Data in Indoor Environments},
  author={Chang, Angel and Dai, Angela and Funkhouser, Thomas and Halber, Maciej and Niessner, Matthias and Savva, Manolis and Song, Shuran and Zeng, Andy and Zhang, Yinda},
  journal={International Conference on 3D Vision (3DV)},
  year={2017}
}

@article{tang2022patchcomplete,
  title={Patchcomplete: Learning multi-resolution patch priors for 3d shape completion on unseen categories},
  author={Rao, Yuchen and Nie, Yinyu and Dai, Angela},
  journal={Advances in Neural Information Processing Systems},
  volume={35},
  pages={34436--34450},
  year={2022}
}

@article{chu2023diffcomplete,
  title={Diffcomplete: Diffusion-based generative 3d shape completion},
  author={Chu, Ruihang and Xie, Enze and Mo, Shentong and Li, Zhenguo and Nie{\ss}ner, Matthias and Fu, Chi-Wing and Jia, Jiaya},
  journal={Advances in neural information processing systems},
  volume={36},
  pages={75951--75966},
  year={2023}
}

@misc{simeoni2025dinov3,
      title={DINOv3}, 
      author={Oriane Siméoni and Huy V. Vo and Maximilian Seitzer and Federico Baldassarre and Maxime Oquab and Cijo Jose and Vasil Khalidov and Marc Szafraniec and Seungeun Yi and Michaël Ramamonjisoa and Francisco Massa and Daniel Haziza and Luca Wehrstedt and Jianyuan Wang and Timothée Darcet and Théo Moutakanni and Leonel Sentana and Claire Roberts and Andrea Vedaldi and Jamie Tolan and John Brandt and Camille Couprie and Julien Mairal and Hervé Jégou and Patrick Labatut and Piotr Bojanowski},
      year={2025},
      eprint={2508.10104},
      archivePrefix={arXiv},
      primaryClass={cs.CV},
      url={https://arxiv.org/abs/2508.10104}, 
}

@article{chang2015shapenet,
  title={Shapenet: An information-rich 3d model repository},
  author={Chang, Angel X and Funkhouser, Thomas and Guibas, Leonidas and Hanrahan, Pat and Huang, Qixing and Li, Zimo and Savarese, Silvio and Savva, Manolis and Song, Shuran and Su, Hao and others},
  journal={arXiv preprint arXiv:1512.03012},
  year={2015}
}

@ARTICLE{adapointr,
author={Yu, Xumin and Rao, Yongming and Wang, Ziyi and Lu, Jiwen and Zhou, Jie},
journal={ IEEE Transactions on Pattern Analysis \& Machine Intelligence },
title={{ AdaPoinTr: Diverse Point Cloud Completion With Adaptive Geometry-Aware Transformers }},
year={2023},
volume={45},
number={12},
ISSN={1939-3539},
pages={14114-14130},
doi={10.1109/TPAMI.2023.3309253},
url = {https://doi.ieeecomputersociety.org/10.1109/TPAMI.2023.3309253},
publisher={IEEE Computer Society},
address={Los Alamitos, CA, USA},
month=dec}

@article{zhang2024voxel,
  title={Voxel mamba: Group-free state space models for point cloud based 3d object detection},
  author={Zhang, Guowen and Fan, Lue and He, Chenhang and Lei, Zhen and Zhang, Zhaoxiang and Zhang, Lei},
  journal={Advances in Neural Information Processing Systems},
  volume={37},
  pages={81489--81509},
  year={2024}
}

@inproceedings{wang2025vggt,
  title={Vggt: Visual geometry grounded transformer},
  author={Wang, Jianyuan and Chen, Minghao and Karaev, Nikita and Vedaldi, Andrea and Rupprecht, Christian and Novotny, David},
  booktitle={2025 IEEE/CVF Conference on Computer Vision and Pattern Recognition (CVPR)},
  pages={5294--5306},
  year={2025},
  organization={IEEE}
}

@inproceedings{caron2021dino,
  title={Emerging properties in self-supervised vision transformers},
  author={Caron, Mathilde and Touvron, Hugo and Misra, Ishan and J{\'e}gou, Herv{\'e} and Mairal, Julien and Bojanowski, Piotr and Joulin, Armand},
  booktitle={2021 IEEE/CVF international conference on computer vision (ICCV)},
  pages={9630--9640},
  year={2021},
  organization={IEEE}
}

@article{
oquab2023dinov2,
title={{DINO}v2: Learning Robust Visual Features without Supervision},
author={Maxime Oquab and Timoth{\'e}e Darcet and Th{\'e}o Moutakanni and Huy V. Vo and Marc Szafraniec and Vasil Khalidov and Pierre Fernandez and Daniel HAZIZA and Francisco Massa and Alaaeldin El-Nouby and Mido Assran and Nicolas Ballas and Wojciech Galuba and Russell Howes and Po-Yao Huang and Shang-Wen Li and Ishan Misra and Michael Rabbat and Vasu Sharma and Gabriel Synnaeve and Hu Xu and Herve Jegou and Julien Mairal and Patrick Labatut and Armand Joulin and Piotr Bojanowski},
journal={Transactions on Machine Learning Research},
issn={2835-8856},
year={2024},
url={https://openreview.net/forum?id=a68SUt6zFt},
note={Featured Certification}
}

@InProceedings{3depn,
author = {Dai, Angela and Ruizhongtai Qi, Charles and Niessner, Matthias},
title = {Shape Completion Using 3D-Encoder-Predictor CNNs and Shape Synthesis},
booktitle = {Proceedings of the IEEE Conference on Computer Vision and Pattern Recognition (CVPR)},
month = {July},
year = {2017}
}

@INPROCEEDINGS{fewshotcompletion,
  author={Wallace, Bram and Hariharan, Bharath},
  booktitle={2019 IEEE/CVF International Conference on Computer Vision (ICCV)}, 
  title={Few-Shot Generalization for Single-Image 3D Reconstruction via Priors}, 
  year={2019},
  volume={},
  number={},
  pages={3817-3826},
  doi={10.1109/ICCV.2019.00392}}

@inproceedings{ifnets,
  title={Implicit functions in feature space for 3d shape reconstruction and completion},
  author={Chibane, Julian and Alldieck, Thiemo and Pons-Moll, Gerard},
  booktitle={2020 IEEE/CVF conference on computer vision and pattern recognition (CVPR)},
  pages={6968--6979},
  year={2020},
  organization={IEEE}
}

@inproceedings{autosdf,
  title={Autosdf: Shape priors for 3d completion, reconstruction and generation},
  author={Mittal, Paritosh and Cheng, Yen-Chi and Singh, Maneesh and Tulsiani, Shubham},
  booktitle={2022 IEEE/CVF Conference on Computer Vision and Pattern Recognition (CVPR)},
  pages={306--315},
  year={2022},
  organization={IEEE}
}

@inproceedings{avetisyan2019scan2cad,
  title={Scan2cad: Learning cad model alignment in rgb-d scans},
  author={Avetisyan, Armen and Dahnert, Manuel and Dai, Angela and Savva, Manolis and Chang, Angel X and Nie{\ss}ner, Matthias},
  booktitle={2019 IEEE/CVF Conference on Computer Vision and Pattern Recognition (CVPR)},
  pages={2609--2618},
  year={2019},
  organization={IEEE}
}

@article{niessner2013real,
  title={Real-time 3D reconstruction at scale using voxel hashing},
  author={Nie{\ss}ner, Matthias and Zollh{\"o}fer, Michael and Izadi, Shahram and Stamminger, Marc},
  journal={ACM Transactions on Graphics (ToG)},
  volume={32},
  number={6},
  pages={1--11},
  year={2013},
  publisher={ACM, New York, NY, USA}
}

@article{dai2017bundlefusion,
  title={Bundlefusion: Real-time globally consistent 3d reconstruction using on-the-fly surface reintegration},
  author={Dai, Angela and Nie{\ss}ner, Matthias and Zollh{\"o}fer, Michael and Izadi, Shahram and Theobalt, Christian},
  journal={ACM Transactions on Graphics (ToG)},
  volume={36},
  number={4},
  pages={1},
  year={2017},
  publisher={ACM, New York, NY, USA}
}

@inproceedings{wang2023autorecon,
  title={Autorecon: Automated 3d object discovery and reconstruction},
  author={Wang, Yuang and He, Xingyi and Peng, Sida and Lin, Haotong and Bao, Hujun and Zhou, Xiaowei},
  booktitle={2023 IEEE/CVF Conference on Computer Vision and Pattern Recognition (CVPR)},
  pages={21382--21391},
  year={2023},
  organization={IEEE}
}

@article{hu2024metric3dv2,
  title={Metric3d v2: A versatile monocular geometric foundation model for zero-shot metric depth and surface normal estimation},
  author={Hu, Mu and Yin, Wei and Zhang, Chi and Cai, Zhipeng and Long, Xiaoxiao and Chen, Hao and Wang, Kaixuan and Yu, Gang and Shen, Chunhua and Shen, Shaojie},
  journal={IEEE Transactions on Pattern Analysis and Machine Intelligence},
  volume={46},
  number={12},
  pages={10579--10596},
  year={2024},
  publisher={IEEE}
}

@inproceedings{wang2025moge,
  title={Moge: Unlocking accurate monocular geometry estimation for open-domain images with optimal training supervision},
  author={Wang, Ruicheng and Xu, Sicheng and Dai, Cassie and Xiang, Jianfeng and Deng, Yu and Tong, Xin and Yang, Jiaolong},
  booktitle={2025 IEEE/CVF Conference on Computer Vision and Pattern Recognition (CVPR)},
  pages={5261--5271},
  year={2025},
  organization={IEEE}
}

@inproceedings{
wang2025mogev2,
title={MoGe-2: Accurate Monocular Geometry with Metric Scale and Sharp Details},
author={Ruicheng Wang and Sicheng Xu and Yue Dong and Yu Deng and Jianfeng Xiang and Zelong Lv and Guangzhong Sun and Xin Tong and Jiaolong Yang},
booktitle={The Thirty-ninth Annual Conference on Neural Information Processing Systems},
year={2026},
url={https://openreview.net/forum?id=16mDq7m2OK}
}

@article{azhari2025dino,
  title={Dino-vo: A feature-based visual odometry leveraging a visual foundation model},
  author={Azhari, Maulana Bisyir and Shim, David Hyunchul},
  journal={IEEE Robotics and Automation Letters},
  year={2025},
  publisher={IEEE}
}

@inproceedings{knaebel2025dino,
  title={Dino in the room: Leveraging 2d foundation models for 3d segmentation},
  author={Knaebel, Karim and Yilmaz, Kadir and De Geus, Daan and Hermans, Alexander and Adrian, David and Linder, Timm and Leibe, Bastian},
  booktitle={2026 International Conference on 3D Vision (3DV)},
  pages={1884--1894},
  year={2026},
  organization={IEEE}
}

@inproceedings{
kim2024openvla,
title={Open{VLA}: An Open-Source Vision-Language-Action Model},
author={Moo Jin Kim and Karl Pertsch and Siddharth Karamcheti and Ted Xiao and Ashwin Balakrishna and Suraj Nair and Rafael Rafailov and Ethan P Foster and Pannag R Sanketi and Quan Vuong and Thomas Kollar and Benjamin Burchfiel and Russ Tedrake and Dorsa Sadigh and Sergey Levine and Percy Liang and Chelsea Finn},
booktitle={8th Annual Conference on Robot Learning},
year={2024},
url={https://openreview.net/forum?id=ZMnD6QZAE6}
}

@INPROCEEDINGS {he2020moco,
author = { He, Kaiming and Fan, Haoqi and Wu, Yuxin and Xie, Saining and Girshick, Ross },
booktitle = { 2020 IEEE/CVF Conference on Computer Vision and Pattern Recognition (CVPR) },
title = {{ Momentum Contrast for Unsupervised Visual Representation Learning }},
year = {2020},
volume = {},
ISSN = {},
pages = {9726-9735},
doi = {10.1109/CVPR42600.2020.00975},
url = {https://doi.ieeecomputersociety.org/10.1109/CVPR42600.2020.00975},
publisher = {IEEE Computer Society},
address = {Los Alamitos, CA, USA},
month =Jun}

@InProceedings{chen2020simclr,
  title = 	 {A Simple Framework for Contrastive Learning of Visual Representations},
  author =       {Chen, Ting and Kornblith, Simon and Norouzi, Mohammad and Hinton, Geoffrey},
  booktitle = 	 {Proceedings of the 37th International Conference on Machine Learning},
  pages = 	 {1597--1607},
  year = 	 {2020},
  editor = 	 {III, Hal Daumé and Singh, Aarti},
  volume = 	 {119},
  series = 	 {Proceedings of Machine Learning Research},
  month = 	 {13--18 Jul},
  publisher =    {PMLR},
  url = 	 {https://proceedings.mlr.press/v119/chen20j.html}
}

@article{grill2020byol,
  title={Bootstrap your own latent-a new approach to self-supervised learning},
  author={Grill, Jean-Bastien and Strub, Florian and Altch{\'e}, Florent and Tallec, Corentin and Richemond, Pierre and Buchatskaya, Elena and Doersch, Carl and Avila Pires, Bernardo and Guo, Zhaohan and Gheshlaghi Azar, Mohammad and others},
  journal={Advances in neural information processing systems},
  volume={33},
  pages={21271--21284},
  year={2020}
}

@incollection{hilbert1935stetige,
  title={{\"U}ber die stetige Abbildung einer Linie auf ein Fl{\"a}chenst{\"u}ck},
  author={Hilbert, David},
  booktitle={Dritter Band: Analysis{\textperiodcentered} Grundlagen der Mathematik{\textperiodcentered} Physik Verschiedenes: Nebst Einer Lebensgeschichte},
  pages={1--2},
  year={1935},
  publisher={Springer}
}

@article{ba2016layer,
  title={Layer normalization},
  author={Ba, Jimmy Lei and Kiros, Jamie Ryan and Hinton, Geoffrey E},
  journal={arXiv preprint arXiv:1607.06450},
  year={2016}
}

@article{gu2023mamba,
  title={Mamba: Linear-time sequence modeling with selective state spaces},
  author={Gu, Albert and Dao, Tri},
  journal={arXiv preprint arXiv:2312.00752},
  year={2023}
}

@article{abdi2010principal,
  title={Principal component analysis},
  author={Abdi, Herv{\'e} and Williams, Lynne J},
  journal={Wiley interdisciplinary reviews: computational statistics},
  volume={2},
  number={4},
  pages={433--459},
  year={2010},
  publisher={Wiley Online Library}
}

@inproceedings{zakeri2026latent,
  title={Latent Uncertainty-Aware Multi-View SDF Scan Completion},
  author={Zakeri, Faezeh and Ruppert, Lukas and Braun, Raphael and Lensch, Hendrik},
  booktitle={Proceedings of the IEEE/CVF Winter Conference on Applications of Computer Vision},
  pages={3556--3566},
  year={2026}
}

@article{schaefer2024sc,
  title={Sc-diff: 3d shape completion with latent diffusion models},
  author={Schaefer, Simon and Galvis, Juan D and Zuo, Xingxing and Leutengger, Stefan},
  journal={arXiv preprint arXiv:2403.12470},
  year={2024}
}

@inproceedings{deitke2023objaverse,
  title={Objaverse: A universe of annotated 3d objects},
  author={Deitke, Matt and Schwenk, Dustin and Salvador, Jordi and Weihs, Luca and Michel, Oscar and VanderBilt, Eli and Schmidt, Ludwig and Ehsani, Kiana and Kembhavi, Aniruddha and Farhadi, Ali},
  booktitle={Proceedings of the IEEE/CVF conference on computer vision and pattern recognition},
  pages={13142--13153},
  year={2023}
}

@inproceedings{
gu2021efficiently,
title={Efficiently Modeling Long Sequences with Structured State Spaces},
author={Albert Gu and Karan Goel and Christopher Re},
booktitle={International Conference on Learning Representations},
year={2022},
url={https://openreview.net/forum?id=uYLFoz1vlAC}
}

@inproceedings{
dao2024transformers,
title={Transformers are {SSM}s: Generalized Models and Efficient Algorithms Through Structured State Space Duality},
author={Tri Dao and Albert Gu},
booktitle={Forty-first International Conference on Machine Learning},
year={2024},
url={https://openreview.net/forum?id=ztn8FCR1td}
}

@inproceedings{
gu2021combining,
title={Combining Recurrent, Convolutional, and Continuous-time Models with Linear State Space Layers},
author={Albert Gu and Isys Johnson and Karan Goel and Khaled Kamal Saab and Tri Dao and Atri Rudra and Christopher Re},
booktitle={Advances in Neural Information Processing Systems},
editor={A. Beygelzimer and Y. Dauphin and P. Liang and J. Wortman Vaughan},
year={2021},
url={https://openreview.net/forum?id=yWd42CWN3c}
}

@inproceedings{zheng2022sdf,
  title={SDF-StyleGAN: implicit SDF-based StyleGAN for 3D shape generation},
  author={Zheng, Xinyang and Liu, Yang and Wang, Pengshuai and Tong, Xin},
  booktitle={Computer Graphics Forum},
  volume={41},
  pages={52--63},
  year={2022},
  organization={Wiley Online Library}
}

@inproceedings{lugmayr2022repaint,
  title={Repaint: Inpainting using denoising diffusion probabilistic models},
  author={Lugmayr, Andreas and Danelljan, Martin and Romero, Andres and Yu, Fisher and Timofte, Radu and Van Gool, Luc},
  booktitle={2022 IEEE/CVF conference on computer vision and pattern recognition (CVPR)},
  pages={11451--11461},
  year={2022},
  organization={IEEE}
}

@inproceedings{yuan2018pcn,
  title={Pcn: Point completion network},
  author={Yuan, Wentao and Khot, Tejas and Held, David and Mertz, Christoph and Hebert, Martial},
  booktitle={2018 international conference on 3D vision (3DV)},
  pages={728--737},
  year={2018},
  organization={IEEE}
}

@inproceedings{yu2021pointr,
  title={Pointr: Diverse point cloud completion with geometry-aware transformers},
  author={Yu, Xumin and Rao, Yongming and Wang, Ziyi and Liu, Zuyan and Lu, Jiwen and Zhou, Jie},
  booktitle={Proceedings of the IEEE/CVF international conference on computer vision},
  pages={12498--12507},
  year={2021}
}

@article{mildenhall2021nerf,
  title={Nerf: Representing scenes as neural radiance fields for view synthesis},
  author={Mildenhall, Ben and Srinivasan, Pratul P and Tancik, Matthew and Barron, Jonathan T and Ramamoorthi, Ravi and Ng, Ren},
  journal={Communications of the ACM},
  volume={65},
  number={1},
  pages={99--106},
  year={2021},
  publisher={ACM, New York, NY, USA}
}

@Article{kerbl3Dgaussians,
      author       = {Kerbl, Bernhard and Kopanas, Georgios and Leimk{\"u}hler, Thomas and Drettakis, George},
      title        = {3D Gaussian Splatting for Real-Time Radiance Field Rendering},
      journal      = {ACM Transactions on Graphics},
      number       = {4},
      volume       = {42},
      month        = {July},
      year         = {2023},
      url          = {https://repo-sam.inria.fr/fungraph/3d-gaussian-splatting/}
}

@inproceedings{schonberger2016structure,
  title={Structure-from-motion revisited},
  author={Schonberger, Johannes L and Frahm, Jan-Michael},
  booktitle={Proceedings of the IEEE conference on computer vision and pattern recognition},
  pages={4104--4113},
  year={2016}
}

@inproceedings{ronneberger2015u,
  title={U-net: Convolutional networks for biomedical image segmentation},
  author={Ronneberger, Olaf and Fischer, Philipp and Brox, Thomas},
  booktitle={International Conference on Medical image computing and computer-assisted intervention},
  pages={234--241},
  year={2015},
  organization={Springer}
}

@inproceedings{kong2026bridgeshape,
  title={BridgeShape: Latent Diffusion Schr{\"o}dinger Bridge for 3D Shape Completion},
  author={Kong, Dequan and Chen, Honghua and Zhu, Zhe and Wei, Mingqiang},
  booktitle={Proceedings of the AAAI Conference on Artificial Intelligence},
  volume={40},
  number={7},
  pages={5726--5734},
  year={2026}
}

@article{liu2023i2sb,
  title={I{$^2$}SB: Image-to-Image Schr{\"o}dinger Bridge},
  author={Liu, Guan-Horng and Vahdat, Arash and Huang, De-An and Theodorou, Evangelos A and Nie, Weili and Anandkumar, Anima},
  journal={arXiv preprint arXiv:2302.05872},
  year={2023},
}

@article{razavi2019generating,
  title={Generating diverse high-fidelity images with vq-vae-2},
  author={Razavi, Ali and Van den Oord, Aaron and Vinyals, Oriol},
  journal={Advances in neural information processing systems},
  volume={32},
  year={2019}
}
\newpage
\appendix

\section{Technical appendices and supplementary material}

\subsection{TSDF-DINO Model}
\begin{figure}[t]
    \centering
    \includegraphics[width=\textwidth]{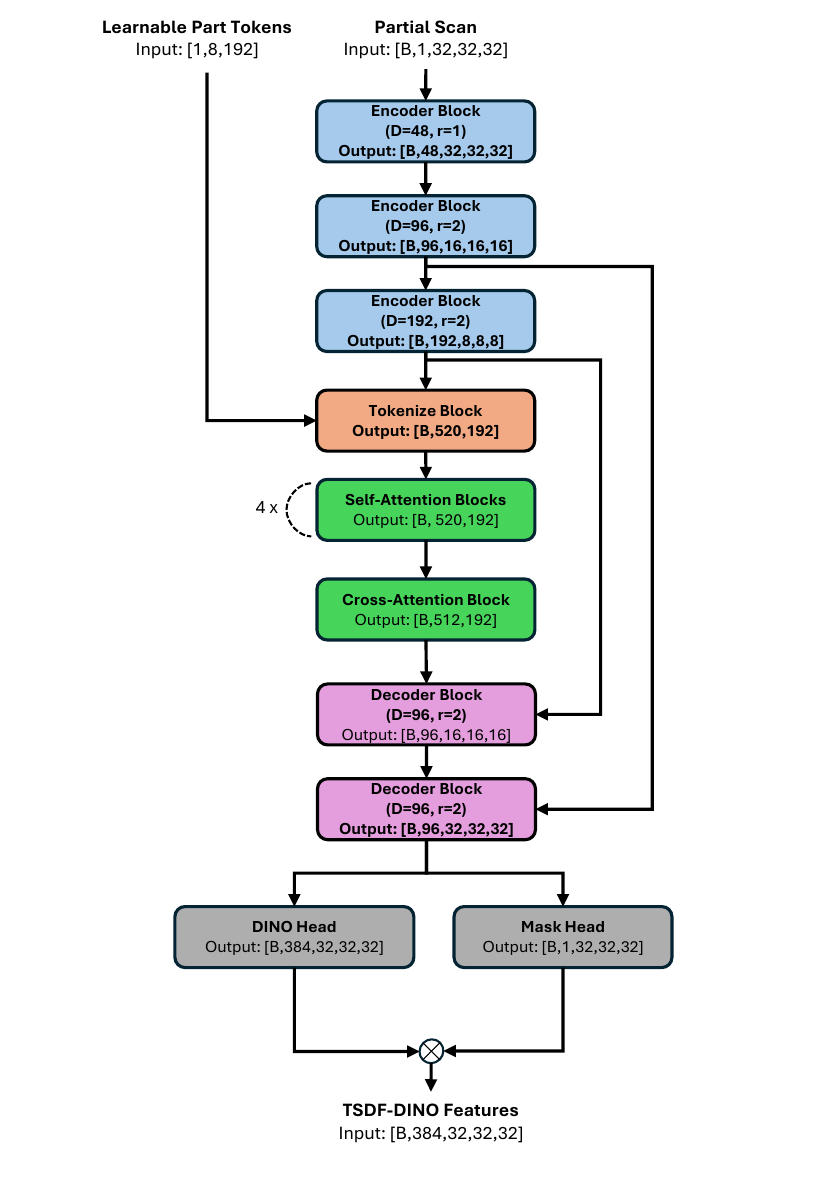}
    \caption{The main architecture of TSDF-DINO model. D denotes feature dimension $r$ denotes downsampling rate.}
    \label{fig:tsdf-dino-architecture}
\end{figure}

\begin{figure}[t]
    \centering
    \includegraphics[width=\textwidth]{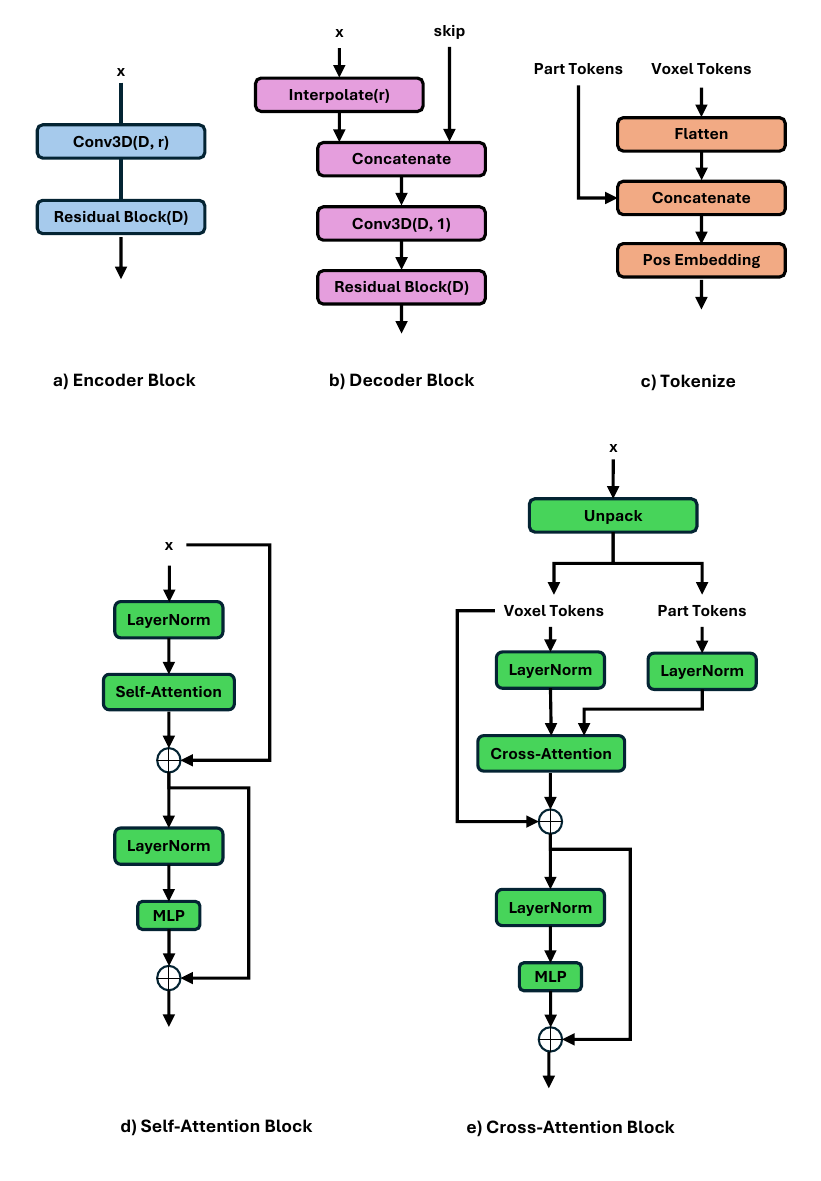}
    \caption{The detailed illustration of model blocks.}
    \label{fig:tsdf-dino-architecture-blocks}
\end{figure}

\begin{figure}[t]
    \centering
    \includegraphics[width=\textwidth]{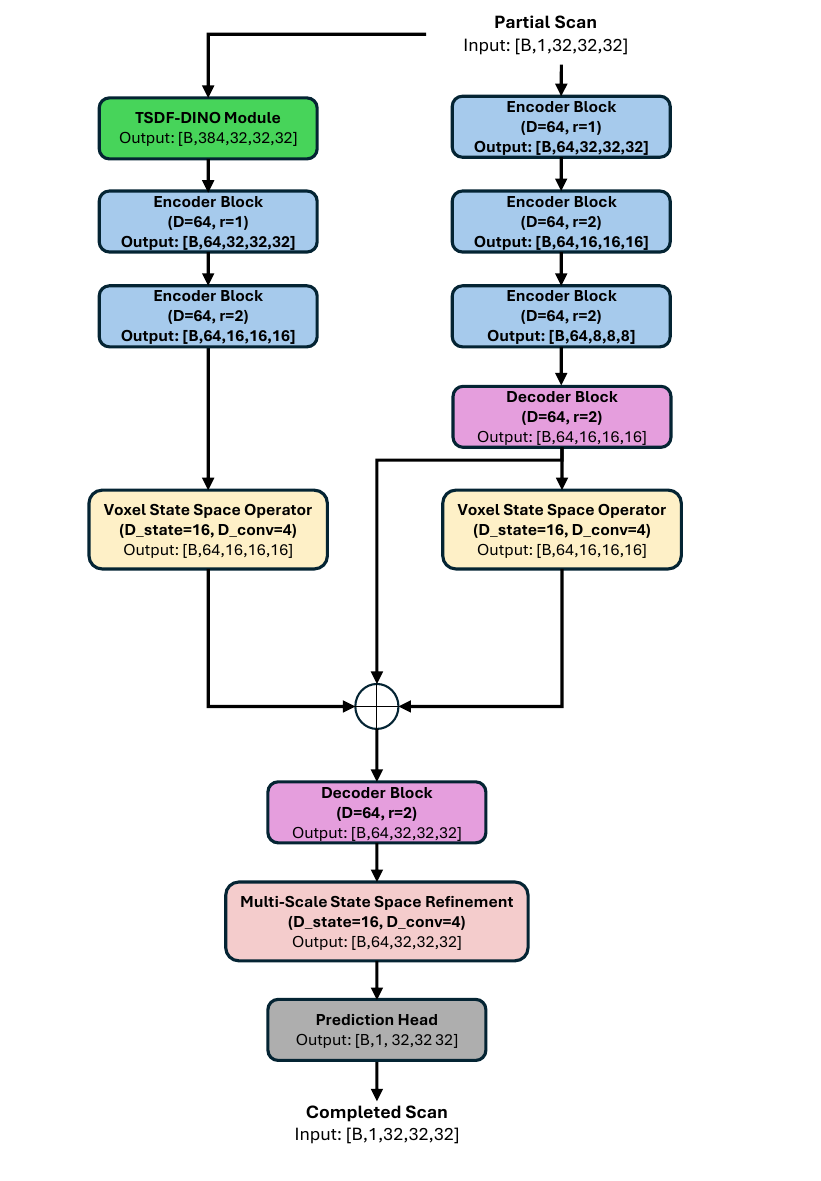}
    \caption{The main overview of shape completion model.}
    \label{fig:architecture-blocks}
\end{figure}

In this section, we provide additional details on our TSDF-DINO model, including the generation of ground-truth data for our student model training, as well as the architectural design. An overview of the TSDF–DINO model and the full shape completion architecture, along with the corresponding hyperparameters, is shown in Figure ~\ref{fig:tsdf-dino-architecture} and Figure~\ref{fig:architecture-blocks}, respectively. Detailed descriptions of the each blocks are provided in Figure ~\ref{fig:tsdf-dino-architecture-blocks}.

\subsubsection{Training Data Preparation}
We precompute voxelized DINOv3~\citep{simeoni2025dinov3} features for each training shape to obtain semantic information aligned with our TSDF representation. We follow the same rendering and fusion setup used in PatchComplete~\citep{tang2022patchcomplete}\footnote{\url{https://github.com/yinyunie/depth_renderer}}, where each ShapeNet~\citep{chang2015shapenet} object is rendered from $V=20$ object-centered multi-view RGB-D images. For each view $v$, we extract dense patch-level visual features from the RGB image $I_v$ using a pretrained DINOv3 backbone(ViT-S/16 v) $\varphi(\cdot)$. The resulting feature map is defined as
\begin{equation}
\mathbf{F}_v = \varphi(I_v), \qquad \mathbf{F}_v \in \mathbb{R}^{C \times H_p \times W_p},
\end{equation}
where $C$ denotes the feature dimension and $(H_p, W_p)$ denotes the spatial patch resolution.

To extract these 2D features and to extend them into 3D, we use the corresponding depth map and camera parameters to associate each patch feature with a 3D location in the canonical object space. Since $\mathbf{F}_v \in \mathbb{R}^{C \times H_p \times W_p}$ contains a feature vector for each patch, we flatten the spatial dimensions into a single index $i \in \{1,\dots,H_pW_p\}$, where each index $i$ corresponds to one patch in the image. We denote the feature of the $i$-th patch in view $v$ as $\mathbf{f}_{v,i} \in \mathbb{R}^C$. 

To compute its 3D location, we consider all valid depth pixels that fall inside patch $i$ and back-project them into the canonical object space, and denote the set of valid pixels as $\Omega_{v,i}$. Let $\mathbf{x}^{\mathrm{can}}_{v}(\mathbf{q}) \in \mathbb{R}^3$ denote the canonical 3D point corresponding to a pixel $\mathbf{q}$ in view $v$. This point is obtained by back-projecting the pixel into 3D camera coordinates using the depth map $D_v$ and camera intrinsics $\mathbf{K}$, and then transforming it into the canonical object coordinate system using the camera extrinsics $\mathbf{R}_v$ and $\mathbf{t}_v$:
\begin{equation}
\mathbf{x}^{\mathrm{can}}_{v}(\mathbf{q})
=
\left( D_v(\mathbf{q}) \, \mathbf{K}^{-1} \tilde{\mathbf{q}} - \mathbf{t}_v \right)\mathbf{R}_v^\top,
\end{equation}
where $\tilde{\mathbf{q}} = [u, w, 1]^\top$ is the homogeneous pixel coordinate. We then define the patch center $\mathbf{p}_{v,i} \in \mathbb{R}^3$ as the average of these 3D points:
\begin{equation}
\mathbf{p}_{v,i}
=
\frac{1}{|\Omega_{v,i}|}
\sum_{\mathbf{q}\in\Omega_{v,i}}
\mathbf{x}^{\mathrm{can}}_{v}(\mathbf{q}).
\end{equation}

We then map each patch center $\mathbf{p}_{v,i}$ into a canonical voxel grid of size $G \times G \times G$.

\begin{equation}
\tilde{\mathbf{p}}_{v,i}
=
(G - 1)\,
\frac{\mathbf{p}_{v,i} - c_{\min}}{c_{\max} - c_{\min}},
\end{equation}

where $c_{\min} = -0.5$ and $c_{\max} = 0.5$ define the bounds of the canonical object space along each axis. Since $\tilde{\mathbf{p}}_{v,i}$ does not generally lie exactly on voxel centers, we distribute its feature $\mathbf{f}_{v,i}$ to the eight neighboring voxels using trilinear interpolation.

Specifically, let $\mathcal{N}(\mathbf{p}_{v,i})$ denote the set of the eight neighboring voxels of $\tilde{\mathbf{p}}_{v,i}$. For each voxel $k \in \mathcal{N}(\mathbf{p}_{v,i})$, we define an interpolation weight $w_{v,i,k}$ based on the relative distance between $\tilde{\mathbf{p}}_{v,i}$ and the voxel center. The weight is given by the standard trilinear interpolation formulation:
\begin{equation}
w_{v,i,k}
=
(1 - |x_i - x_k|)\,(1 - |y_i - y_k|)\,(1 - |z_i - z_k|), 
\end{equation}
where $(x_i, y_i, z_i)$ correspond to the coordinates of $\tilde{\mathbf{p}}_{v,i}$ and neighbor voxel $k$, respectively.

Using these weights, we accumulate features from all patches into each voxel. The accumulated feature sum and weight sum for voxel $k$ in view $v$ are defined as
\begin{equation}
\mathbf{S}_v(k) = \sum_{i:\,k \in \mathcal{N}(\tilde{\mathbf{p}}_{v,i})} w_{v,i,k} \, \mathbf{f}_{v,i}, \qquad
W_v(k) = \sum_{i:\,k \in \mathcal{N}(\tilde{\mathbf{p}}_{v,i})} w_{v,i,k}.
\label{eq:weight_sum}
\end{equation}
The per-view voxel feature is then obtained by normalization:
\begin{equation}
\mathbf{G}_v(k) = \frac{\mathbf{S}_v(k)}{W_v(k) + \epsilon}.
\end{equation}

To ensure that only valid object regions contribute to the supervision, we further filter voxel features using the ground-truth TSDF. Let $T(\mathbf{x})$ denote the ground-truth TSDF volume. We construct a binary mask
\begin{equation}
M(k)=\mathbb{I}\big[T(k)\leq 0\big],
\end{equation}
which keeps only voxels inside the object surface. The filtered per-view features are thus
\begin{equation}
\tilde{\mathbf{G}}_v(k)=M(k)\,\mathbf{G}_v(k),
\qquad
\tilde{W}_v(k)=M(k)\,W_v(k).
\end{equation}

Finally, we fuse all $V$ views into a single ground-truth 3D DINO volume by weighted averaging across views:
\begin{equation}
\mathbf{F}^{\text{GT}}(k)
=
\frac{\sum_{v=1}^{V}\tilde{W}_v(k)\,\tilde{\mathbf{G}}_v(k)}
{\sum_{v=1}^{V}\tilde{W}_v(k)+\epsilon}.
\end{equation}

\begin{figure}[t]
    \centering
    \includegraphics[width=\textwidth]{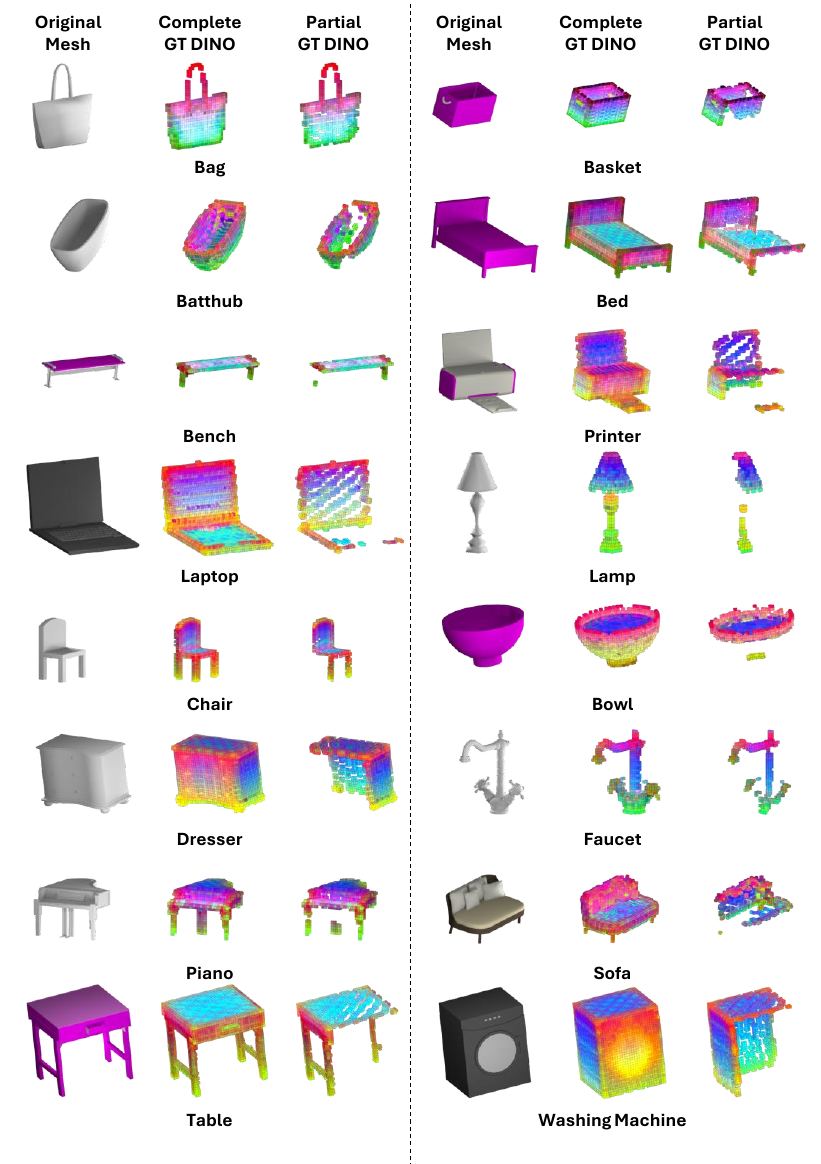}
    \caption{The generated ground truth data for TSDF-DINO model.}
    \label{fig:gt-dino}
\end{figure}

To generate incomplete DINO supervision consistent with partial TSDF inputs, for a selected input view $v$, we compute a visibility (coverage) mask in the voxel grid using its depth map $D_v$ and the same trilinear voxelization procedure. Specifically, we reuse the per-view accumulated weights $W_v(k)$ as defined in Eq.~\ref{eq:weight_sum}, which measure the contribution of projected geometry to voxel $k$. The coverage mask and the resulting incomplete DINO target are defined as
\begin{equation}
M_v^{\mathrm{cov}}(k) = \mathbb{I}\big[ W_v(k) > 0 \big], \quad
\mathbf{F}^{\mathrm{inc}}_v(k) = M_v^{\mathrm{cov}}(k)\,\mathbf{F}^{\mathrm{GT}}(k).
\end{equation}
The resulting partial 3D DINO volume is spatially aligned with the corresponding incomplete TSDF input while preserving the semantic consistency of the multi-view fused representation. Examples of fully fused DINO data and partial DINO data, along with their corresponding colored meshes, are shown in Figure \ref{fig:gt-dino}.

\subsubsection{Architecture Overview}
\paragraph{Encoding and Feature Extraction}
Our TSDF-DINO model follows a hybrid design that combines U-Net structures with attention blocks. Given a partial TSDF $X \in \mathbb{R}^{1 \times 32 \times 32 \times 32}$, we first extract voxel features using 3D CNNs. These coarse voxel features are then projected into a token space, forming a sequence of \textbf{voxel tokens}. In addition, we introduce a set of learnable part tokens that represent latent semantic components of the shape. The voxel and part tokens are concatenated and then put through positional embeddings, forming the input to a transformer backbone.
\paragraph{Tokenization and Attention Blocks}
Inspired by VGGT~\cite{wang2025vggt}, our transformer uses token-level attention to combine local structure with global context.In our voxel setting, the joint self-attention blocks act as a global attention module over the full set of voxel and part tokens, enabling information exchange across the entire $8^3$ volume. After this global reasoning stage, we apply a part-aware cross-attention block in which voxel tokens query the learned part tokens. The resulting voxel tokens therefore encode both global 3D context and part-conditioned semantic structure.
\paragraph{Dense Reconstruction and Prediction.}
The refined voxel tokens are reshaped back into a dense $8^3$ grid and passed through a hierarchical 3D decoder with skip connections to recover spatial detail at $32^3$ resolution. The final feature volume is used to predict voxel-aligned DINO features. We use a mask head that predicts per-voxel confidence, which is used to generate the final output. This design enables the model to focus on reliable regions of the shape while suppressing uncertain areas. 

\begin{table}[H]
\centering
\caption{Feature alignment results of our method on ShapeNet~\cite{chang2015shapenet}. Cosine similarity ($\uparrow \times 10^{2}$) and MSE ($\downarrow \times 10^{-2}$) are reported for unseen and seen categories.}
\footnotesize
\setlength{\tabcolsep}{3.5pt}
\resizebox{\columnwidth}{!}{
\begin{tabular}{c|ccccccccc}
\toprule
\multicolumn{10}{c}{\textbf{Unseen Categories}} \\
\midrule
\textbf{Avg.} 
& Bag & Basket & Bathtub & Bed & Bench & Lamp & Laptop & Printer &  \\
\midrule
\textbf{87.0 / 2.98}
& 82.5 / 4.04 
& 82.5 / 4.09 
& 89.8 / 2.36 
& 87.6 / 2.83 
& 88.3 / 2.63 
& 85.0 / 3.57 
& 87.2 / 2.87 
& 89.1 / 2.47 
&  \\

\midrule\midrule

\multicolumn{10}{c}{\textbf{Seen Categories}} \\
\midrule
\textbf{Avg.} 
& Trash Bin & Bookshelf & Bowl & Cabinet & Chair & Keyboard & Dishwasher & Display & Faucet \\
\midrule
\textbf{89.9 / 2.37}
& 88.7 / 2.69 
& 88.0 / 2.69 
& 92.2 / 1.89 
& 90.3 / 2.38 
& 82.1 / 3.99 
& 94.7 / 1.19 
& 93.4 / 1.54 
& 90.6 / 2.41 
& 90.9 / 2.19 \\

\midrule

\multicolumn{1}{c}{} 
& File Cabinet & Guitar & Microwave & Piano & Pot & Sofa & Stove & Table & Washing Machine \\
\midrule
\multicolumn{1}{c}{} 
& 91.0 / 2.07 
& 89.3 / 2.80 
& 92.8 / 1.72 
& 90.0 / 2.22 
& 88.7 / 2.78 
& 88.2 / 2.69 
& 89.2 / 2.50 
& 88.4 / 2.56 
& 89.8 / 2.41 \\

\bottomrule
\end{tabular}
}
\label{tab:tsdf-dino-qual}
\end{table}

\subsubsection{Similarity Results}
\begin{figure}[t]
    \centering
    \includegraphics[width=\textwidth]{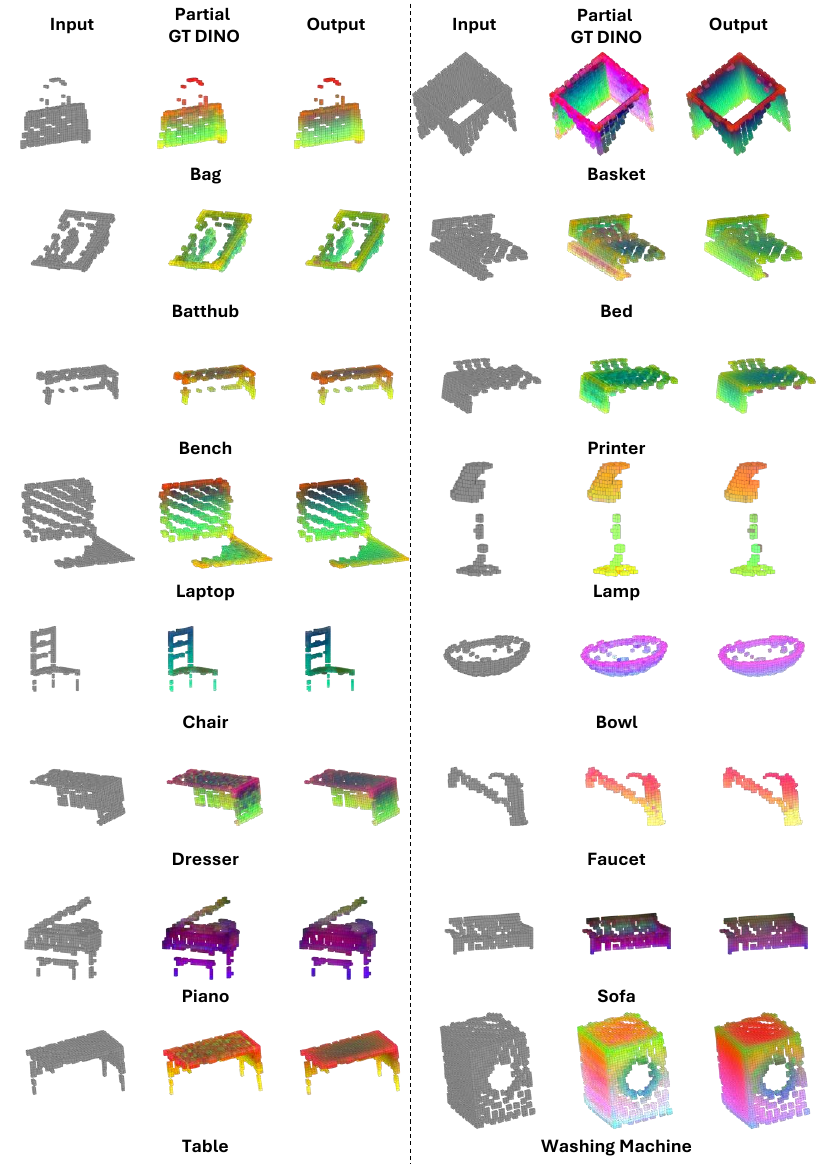}
    \caption{Additional qualitative results of our TSDF-DINO model.}
    \label{fig:output-dino}
\end{figure}

Table~\ref{tab:tsdf-dino-qual} presents the feature alignment results of our student model on ShapeNet. We report cosine similarity and mean squared error (MSE), highlighting performance across both seen and unseen categories. Strong performance on unseen categories indicates that our TSDF–DINO model can infer similar part-level information without being trained on every category. We also provide additional qualitative results in Figure~\ref{fig:output-dino}, showing the uncolored input voxels, ground-truth DINO features, and our model’s outputs. We perform PCA on both the ground truth and our predictions to demonstrate similarity between them.

\subsection{Preliminaries: State Space Models for Voxel Sequences}
\label{sec:ssm_preliminaries}
State space models (SSMs)~\citep{gu2023mamba,gu2021efficiently, dao2024transformers, gu2021combining} provide an efficient alternative to self-attention for modeling long-range dependencies in sequential data. A continuous-time SSM maps an input signal $x(t)$ to an output signal $y(t)$ through a latent state $h(t)$:
\begin{equation}
h'(t) = A h(t) + B x(t), \qquad
y(t) = C h(t) + D x(t),
\end{equation}
where $A\in \mathbb{R}^{N \times N}$, $B\in \mathbb{R}^{N \times 1}$, $C\in \mathbb{R}^{1 \times N}$ are learnable parameters and $D\in \mathbb{R}^{1}$ denotes a residual connection. After discretization, the model can be written as 
\begin{equation}
h_k = \bar{A} h_{k-1} + \bar{B} x_k, \qquad
y_k = \bar{C} h_k + \bar{D} x_k .
\end{equation}
where $\bar{A} = \exp(\Delta A)$ and $\bar{B} = (\Delta A)^{-1}(\exp(\Delta A) - I)\Delta B$ are obtained via zero-order hold (ZOH) discretization with step size $\Delta$. To apply SSMs to 3D data, voxel features are serialized into a one-dimensional sequence. However, naive serialization may separate neighboring voxels in the resulting sequence and reduce spatial locality. Voxel Mamba~\cite{zhang2024voxel} addresses this issue by using a group-free formulation, where the full voxel space is ordered into a single sequence using a Hilbert curve~\citep{hilbert1935stetige}. This preserves local 3D topology better and allows global voxel interactions without explicit window grouping. In our work, we follow this principle in our Voxel State Operators.

\begin{table}[H]
\centering
\caption{Quantitative shape completion results on known ShapeNet categories~\cite{3depn} ($l_1$ error $\downarrow$).}
\footnotesize
\setlength{\tabcolsep}{3pt}
\resizebox{\columnwidth}{!}{
\begin{tabular}{l|c|cccccccc}
\toprule
\textbf{Method} & \textbf{Avg.} & Chair & Table & Sofa & Lamp & Plane & Car & Dresser & Boat \\
\midrule
3D-EPN~\cite{3depn}
& 0.374 & 0.418 & 0.377 & 0.392 & 0.388 & 0.421 & 0.259 & 0.381 & 0.356 \\

SDF-StyleGAN~\cite{zheng2022sdf}
& 0.278 & 0.321 & 0.256 & 0.289 & 0.280 & 0.295 & 0.224 & 0.273 & 0.282 \\

RePaint-3D~\cite{lugmayr2022repaint}
& 0.266 & 0.289 & 0.264 & 0.266 & 0.268 & 0.302 & 0.214 & 0.285 & 0.243 \\

AutoSDF~\cite{autosdf}
& 0.217 & 0.201 & 0.258 & 0.226 & 0.275 & 0.184 & 0.187 & 0.248 & 0.157 \\

PatchComplete~\cite{tang2022patchcomplete}
& 0.088 & 0.134 & 0.095 & 0.084 & 0.087 & 0.061 & 0.053 & 0.134 & 0.058 \\

DiffComplete~\cite{chu2023diffcomplete}
& 0.053 & 0.070 & 0.073 & 0.061 & 0.059 & 0.015 & 0.025 & 0.086 & 0.031 \\

SC-Diff~\cite{schaefer2024sc}
& 0.055 & 0.082 & 0.074 & 0.070 & 0.047 & 0.015 & 0.032 & 0.095 & 0.028 \\

BridgeShape~\cite{kong2026bridgeshape}
& 0.039 & 0.055 & 0.059 & 0.047 & 0.038 & 0.012 & 0.023 & \textbf{0.055} & 0.022 \\
\midrule
\textbf{Ours}
& \textbf{0.036} & \textbf{0.045} & \textbf{0.052} & \textbf{0.042}
& \textbf{0.035} & \textbf{0.011} & \textbf{0.021} & 0.060 & \textbf{0.019} \\
\bottomrule
\end{tabular}
}
\label{tab:known_categories}
\end{table}

\begin{figure}[H]
    \centering
    \includegraphics[width=\linewidth]{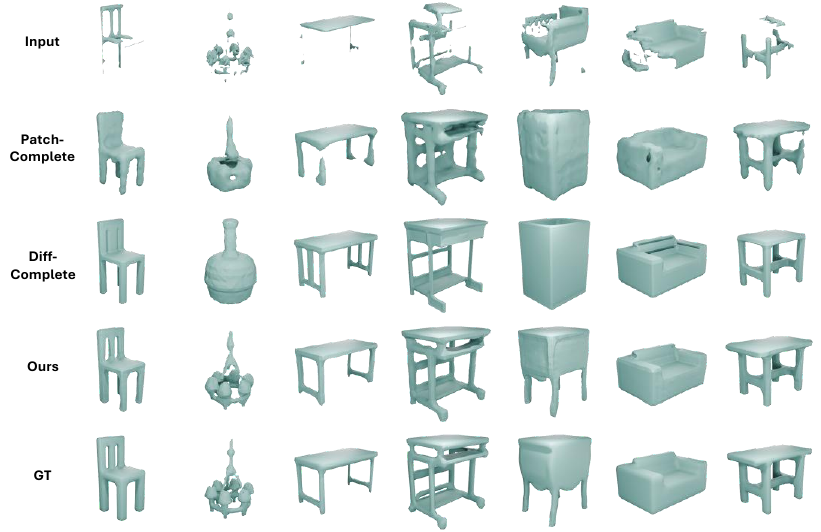}
    \caption{Shape completion results on known categories on 3D-EPN dataset.}
    \label{fig:known-category-results}
\end{figure}

\subsection{Additional Experiments}

\subsubsection{Shape Completion on Unknown Categories}
\begin{figure}[t]
    \centering
    \includegraphics[width=\linewidth]{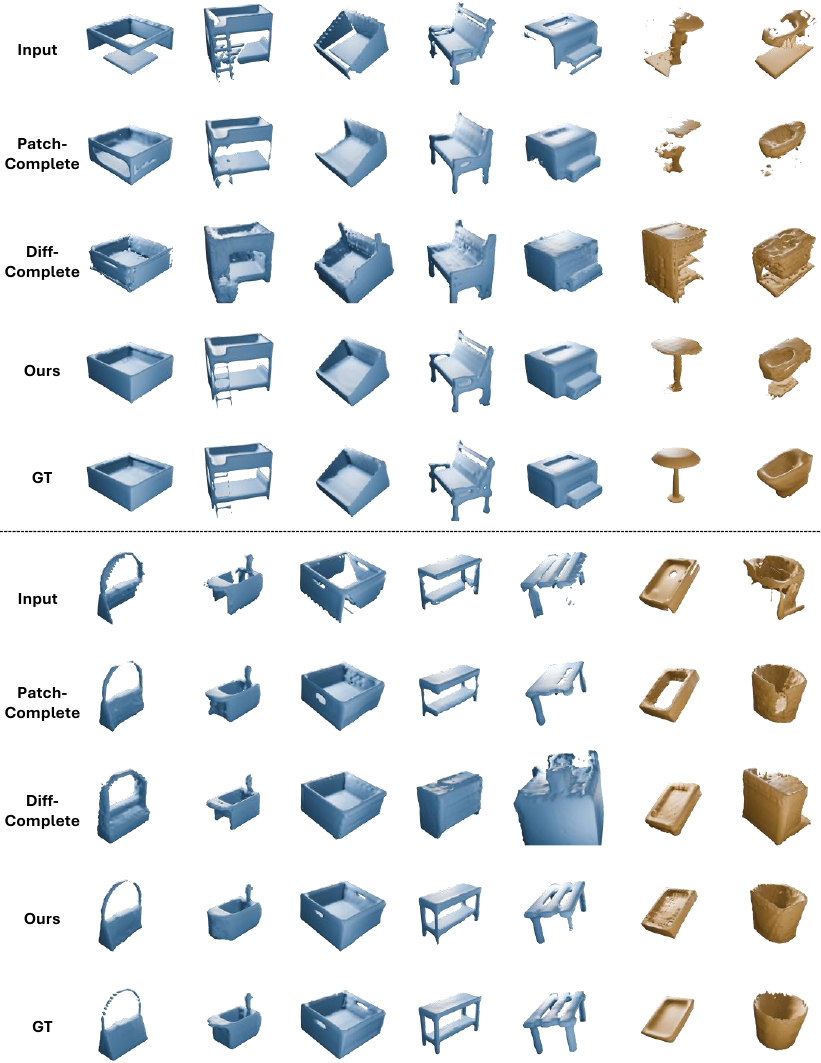}
    \caption{Additional shape completion results  on both synthetic (blue) and real-world (yellow) objects from
entirely unseen categories.}
    \label{fig:results-supplementary}
\end{figure}

We provide additional qualitative results on unseen categories in Figure~\ref{fig:results-supplementary}, extending the results presented in the main paper.

\subsubsection{Shape Completion on Known Categories}
In addition to our focus on unseen-category generalization, we also evaluate our method on known ShapeNet categories using the dataset introduced by 3D-EPN~\cite{3depn}. This dataset differs from the PatchComplete setup~\cite{tang2022patchcomplete} in both data generation and problem formulation. In contrast to PatchComplete, which targets category-agnostic generalization, 3D-EPN benchmark evaluates completion within a fixed set of known categories.

For this benchmark, we train our model in a category-wise manner on the 3D-EPN dataset with same training parameters. Importantly, we use the same TSDF-DINO model learned on the PatchComplete ShapeNet dataset, demonstrating its adaptability across different data distributions. Despite this dataset shift, our method still achieves state-of-the-art performance. As shown in Table~\ref{tab:known_categories}, our method achieves the lowest average $l_1$ reconstruction error among the compared methods. These results indicate that our approach is not only effective for unseen-category generalization but also highly competitive in standard category-specific shape completion settings, while preserving representations learned from a different dataset.

We also present qualitative results in Figure~\ref{fig:known-category-results}. Our method generally mitigates overfitting across diverse shapes. As shown in first example our model accurately preserves details of the individual bulbs in a chandelier. In contrast, PatchComplete tends to underfit the task, while DiffComplete shows signs of significant overfitting In the final bench example, DiffComplete produces a reasonable reconstruction but fails to capture the correct surface extent, resulting in a shortened structure. These observations demonstrate that our model maintains a strong balance between underfitting and overfitting in shape completion.

\newpage

\subsubsection{Offline Preparation Cost}
\label{sec:offline_cost}

Table~\ref{tab:offline_preparation_cost} reports the additional
offline preparation cost under our experimental setup.
Teacher feature volumes are constructed once, cached and reused
during student pretraining. The pretrained student weights then
initialize the semantic branch for downstream shape-completion
training. At inference, the completion model uses the student
without DINOv3 execution or teacher-volume construction.

\begin{table}[t]
\centering
\caption{Additional offline preparation cost before
shape-completion training.}
\label{tab:offline_preparation_cost}
\small
\setlength{\tabcolsep}{5pt}
\begin{adjustbox}{max width=\linewidth}
\begin{tabular}{llcc}
\toprule
Stage
& Setting
& \makecell{Time\\(min)}
& \makecell{Peak GPU memory\\(GB)} \\
\midrule
3D DINO teacher construction
& \makecell[l]{DINO feature extraction\\and $32^3$ voxel fusion}
& 50 & 1,633 \\

TSDF-DINO student pretraining
& Batch size 4
& 120 & 7,923 \\
\midrule
Total model preparation
& \makecell[l]{Offline target construction\\and student pretraining}
& 170 & 7,923 \\
\bottomrule
\end{tabular}
\end{adjustbox}
\end{table}

\subsubsection{Impact of Using Ground-Truth DINO Features}
We compare our model with a variant that directly uses DINOv3 features instead of the learned TSDF–DINO student features, as shown in Table~\ref{tab:dino_student_limitation}. Directly injecting ground-truth DINO features leads to improved performance, indicating that the quality of the distilled features remains a limiting factor. However, we do not adopt this setting in our main model, as it requires image-based features at inference time. Our goal is to maintain a \textbf{TSDF-only input pipeline}, which is more practical in real-world scenarios where only depth observations may be available. This suggests that further improving the TSDF–DINO distillation process could yield additional gains in shape completion performance.

\begin{table}[t]
\centering

\begin{minipage}[t]{0.48\linewidth}
\vspace{0pt}
\centering
\captionof{table}{Comparison between our model and direct DINO
feature injection on unseen categories.}
\label{tab:dino_student_limitation}
\small
\setlength{\tabcolsep}{6pt}
\begin{adjustbox}{max width=\linewidth}
\begin{tabular}{lcc}
\toprule
Model & CD $\downarrow$ & IoU $\uparrow$ \\
\midrule
Ours       & 3.64 & 71.1 \\
DINO input & \textbf{3.27} & \textbf{75.4} \\
\bottomrule
\end{tabular}
\end{adjustbox}
\end{minipage}
\hfill
\begin{minipage}[t]{0.48\linewidth}
\vspace{0pt}
\centering
\captionof{table}{Branch-combination ablation on unseen categories.}
\label{tab:msm_branches}
\small
\setlength{\tabcolsep}{6pt}
\begin{adjustbox}{max width=\linewidth}
\begin{tabular}{lcc}
\toprule
Active branches & CD $\downarrow$ & IoU (\%) $\uparrow$ \\
\midrule
$F_{32}$             & 3.86 & 69.1 \\
$F_8$                & 3.73 & 70.0 \\
$F_4$                & 3.76 & 69.6 \\
$F_8 + F_4$          & 3.75 & 69.6 \\
$F_{32} + F_4$       & 3.75 & 69.8 \\
$F_{32} + F_8$       & 3.67 & 70.3 \\
$F_{32} + F_8 + F_4$ & \textbf{3.64} & \textbf{71.1} \\
\bottomrule
\end{tabular}
\end{adjustbox}
\end{minipage}

\par\vspace{12pt}

\begin{minipage}[t]{\linewidth}
\centering
\captionof{table}{Comparison of refinement architectures
on unseen categories.}
\label{tab:msm_alternatives}
\small
\setlength{\tabcolsep}{5pt}
\begin{adjustbox}{max width=\linewidth}
\begin{tabular}{lccccc}
\toprule
Refinement architecture
& CD $\downarrow$
& IoU (\%) $\uparrow$
& \makecell{Params $\downarrow$\\(M)}
& \makecell{Memory $\downarrow$\\(GB)}
& \makecell{Time $\downarrow$\\(ms)} \\
\midrule
Multi-scale dilated Conv3D
& 3.72 & 70.3 & 21.24 & \textbf{0.237} & 31.70 \\

Hierarchical window attention
& 3.79 & 69.5 & 21.29 & 0.287 & 31.81 \\

Sparse window transformer
& 3.67 & 70.7 & \textbf{21.18} & 0.438 & 36.78 \\

Multi-Scale Voxel Mamba (ours)
& \textbf{3.64} & \textbf{71.1} & 25.97
& 0.291 & \textbf{28.67} \\
\bottomrule
\end{tabular}
\end{adjustbox}
\end{minipage}

\end{table}

\subsubsection{Multi-Scale Voxel Mamba Ablations}
\label{sec:msm_ablation}

We evaluate the contribution of the full-grid and chunk-wise branches and compare our MSM module with alternative refinement architectures. All experiments use unseen ShapeNet categories at $32^3$ resolution under the protocol of Table~\ref{tab:synthetic_unseen}. We keep the remaining architecture and training protocol unchanged.

\paragraph{Contribution of individual branches.}
We denote the full-grid branch by $F_{32}$ and the branches using $4^3$ and $8^3$ voxel chunks by $F_4$ and $F_8$, respectively. The full-grid branch processes individual voxel tokens, while the chunk branches embed local regions into tokens and model interactions across the chunk grid. Table~\ref{tab:msm_branches} compares individual branches and their combinations. Combining all branches achieves the lowest CD and highest IoU, supporting the benefit of regional context complementary to full-grid voxel reasoning.

\paragraph{Comparison with alternative refinement architectures.}
We replace our MSM module with multi-scale dilated Conv3D, hierarchical shifted-window attention, and a sparse-window transformer. For each alternative, we report its strongest tested configuration. The convolutional variant uses two residual stages with parallel depthwise dilated convolutions and an expansion ratio of two. Hierarchical attention uses four heads and local $4^3$ windows with a shift of two voxels. The sparse transformer uses four heads and two alternating regular and shifted window blocks. Table~\ref{tab:msm_alternatives} reports completion accuracy, model parameters, peak inference memory, and per-sample latency. Our module achieves the best completion accuracy and lowest latency among the tested alternatives, with a larger parameter count. Its peak memory is close to hierarchical window attention and lower than the sparse transformer, while dilated Conv3D uses less memory. These results support our multi-scale refinement formulation as an effective choice for accurate completion with fast inference.

\begin{figure}[H]
    \centering
    \includegraphics[width=0.9\linewidth]{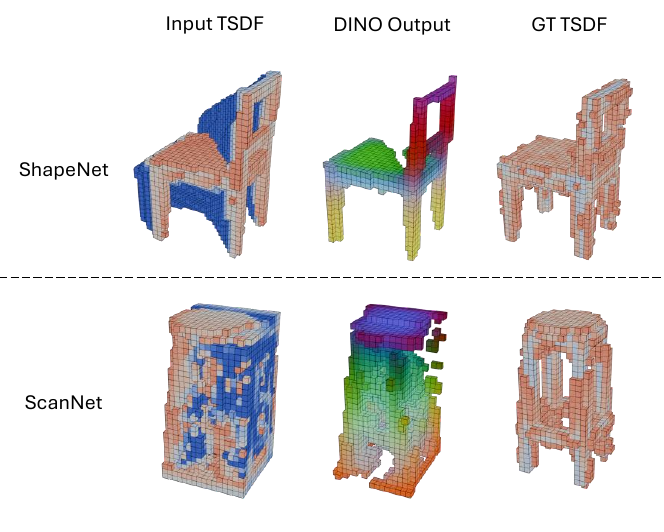}
    \caption{The effect of occlussion between synthetic and real data on our TSDF-DINO model.}
    \label{fig:limitation}
\end{figure}

\section{Evaluation Variablity and Error Bars}
\label{sec:errorbars}
The evaluations are conducted on large-scale test sets with 1325 ShapeNet~\citep{chang2015shapenet} models(with 4 partial scans for each model) and 1191 ScanNet~\citep{dai2017scannet} samples across multiple unseen categories, providing statistically reliable and stable results. Table \ref{tab:synthetic_unseen} and Table \ref{tab:real_unseen} report category-wise error bars on ShapeNet and ScanNet, respectively. Each method is evaluated over $n=2$ independent runs. For SC-Diff~\citep{schaefer2024sc} and BridgeShape~\citep{kong2026bridgeshape} we use their published results because complete implementations were unavailable during our experiments. Their unreported error bars are marked as --.





\section{Licenses and Terms of Use}

This work uses the DINOv3 pretrained models available at \url{https://github.com/facebookresearch/dinov3}, the ShapeNet dataset available at \url{https://shapenet.org/}, the depth rendering implementation available at \url{https://github.com/yinyunie/depth_renderer}, and the ScanNet dataset and code available at \url{https://github.com/ScanNet/ScanNet}. Access to ShapeNet, ScanNet, and DINOv3 was obtained through the official registration and agreement procedures, and the corresponding terms of use were accepted.

\section{Limitations}

Despite its strong performance, our model still has room for improvement. As indicated in Table~\ref{tab:dino_student_limitation}, there remains a gap between the learned TSDF–DINO features and directly injected DINO features, suggesting that the current distillation into voxel space is not yet optimal. Improving this alignment could further enhance performance.

There remains a significant domain gap between synthetic and real-world data. Compared to ShapeNet~\citep{chang2015shapenet}, ScanNet~\citep{dai2017scannet} inputs show substantially more unordered scanning artifacts. As shown in Figure~\ref{fig:limitation}, this results in degraded semantic feature quality. Since our model is trained primarily on synthetic data, it does not explicitly account for this domain shift and lacks mechanisms to robustly handle noisy real-world inputs.

\section{Broader Impact}
By improving generalization to unseen data while using fewer computational resources, our approach has the potential to support real-time reconstruction on limited hardware. This is important for applications in robotics and virtual systems operating in dynamic spaces. Furthermore, improved reconstruction quality can enable more accurate and semantically consistent 3D representations, benefiting tasks such as object understanding, manipulation, and scene interaction in robotics, vision systems, and virtual environments.

However, these advancements also introduce potential risks. The ability to efficiently reconstruct detailed objects could be misused to recreate private spaces or personal belongings without consent, raising privacy concerns. There are also potential economic implications, as increased automation in reconstruction pipelines may reduce reliance on manual 3D modeling in some scenarios.

Overall, while our method advances scalable 3D reconstruction, it is important for future work to incorporate safeguards, such as consent-aware data practices and clear usage guidelines, to minimize misuse while preserving its potential benefits.

\clearpage

\end{document}